\documentclass{article}

    \PassOptionsToPackage{numbers, compress}{natbib}
\usepackage[preprint]{neurips_2026}

\usepackage[utf8]{inputenc} 
\usepackage[T1]{fontenc}    
\usepackage{hyperref}       
\usepackage{url}            
\usepackage{booktabs}       
\usepackage{amsfonts}       
\usepackage{nicefrac}       
\usepackage{microtype}      
\usepackage{xcolor}         
\usepackage{amsmath}
\usepackage{mathtools}
\usepackage{multirow}
\usepackage{tcolorbox}
\tcbuselibrary{breakable, skins}
\usepackage{bm}
\usepackage{bbm}
\usepackage{tikz}
\usepackage{wrapfig}
\usepackage{listofitems}
\usepackage{enumitem}
\usepackage{cleveref}
\usepackage{todonotes}
\usepackage{listings}
\usepackage{xcolor}
\usepackage{makecell}
\usepackage{markdown}
\usepackage{subcaption}
\newlength{\itemizelength} 
\newcommand{\bfx}{{\mathbf{x}}}
\newcommand{\R}{{\mathbb{R}}}
\newcommand{\eps}{{\epsilon}}
\newcommand{\mlpx}{{\mathbf{MLP}_{\mathbf{X}}}}
\newcommand{\mlpt}{{\mathbf{MLP}_T}}
\newcommand{\mlpy}{{\mathbf{MLP}_Y}}

\tikzset{
    sum/.style={
        circle,
        draw=black,
        thick,
        minimum size=0.5cm, 
        inner sep=0pt,
        path picture={
            \draw[thick, black] 
            (path picture bounding box.north) -- (path picture bounding box.south);
            \draw[thick, black] 
            (path picture bounding box.east) -- (path picture bounding box.west);
        }
    }
}

\usetikzlibrary{positioning, calc, arrows.meta}

\title{Causal Foundation Models with \\ Continuous Treatments}

\author{%
  Christopher Stith\thanks{Equal Contribution} \\
    Layer 6 AI\\
    \small\texttt{christopher@layer6.ai}
    \And
    Medha Barath$^*$ \\
    University of Toronto\\
    \small\texttt{medha.barath@mail.utoronto.ca}
    \And
        Vahid Balazadeh \\
    University of Toronto\\
    Vector Institute\\
    \small\texttt{vahid@cs.toronto.edu}
    \AND
  Jesse C. Cresswell \\
    Layer 6 AI\\
    \small\texttt{jesse@layer6.ai}
    \And
        Rahul G. Krishnan \\
    University of Toronto\\
    Vector Institute\\
    \small\texttt{rahulgk@cs.toronto.edu}
    }

\begin{document}

\maketitle

\vspace{-15pt}
\begin{abstract}
Causal inference, estimating causal effects from observational data, is a fundamental tool in many disciplines. Of particular importance across a variety of domains is the \emph{continuous treatment} setting, where the variable of intervention has a continuous range. This setting is far less explored and represents a substantial shift from the binary treatment setting, with models needing to represent effects across a continuum of treatment values. In this paper, we present the first \textit{causal foundation model} for the continuous treatment setting. Our model meta-learns the ability to predict causal effects across a wide variety of unseen tasks without additional training or fine-tuning. First, we design a novel prior over data-generating processes with continuous treatment variables in order to generate a rich causal training corpus. We then train a transformer to reconstruct individual treatment-response curves given only observational data, leveraging in-context learning to amortize expensive Bayesian posterior inference. Our model achieves state-of-the-art performance on individual treatment-response curve reconstruction tasks compared to causal models which are trained specifically for those tasks. Inference code (including trained model weights) can be found at \href{https://github.com/layer6ai-labs/CCPFN-inference}{\tt{github.com/layer6ai-labs/CCPFN-inference}}.
\end{abstract}

\vspace{-14pt}
\begin{figure*}[h]
    \centering
    \includegraphics[width=0.8\textwidth]{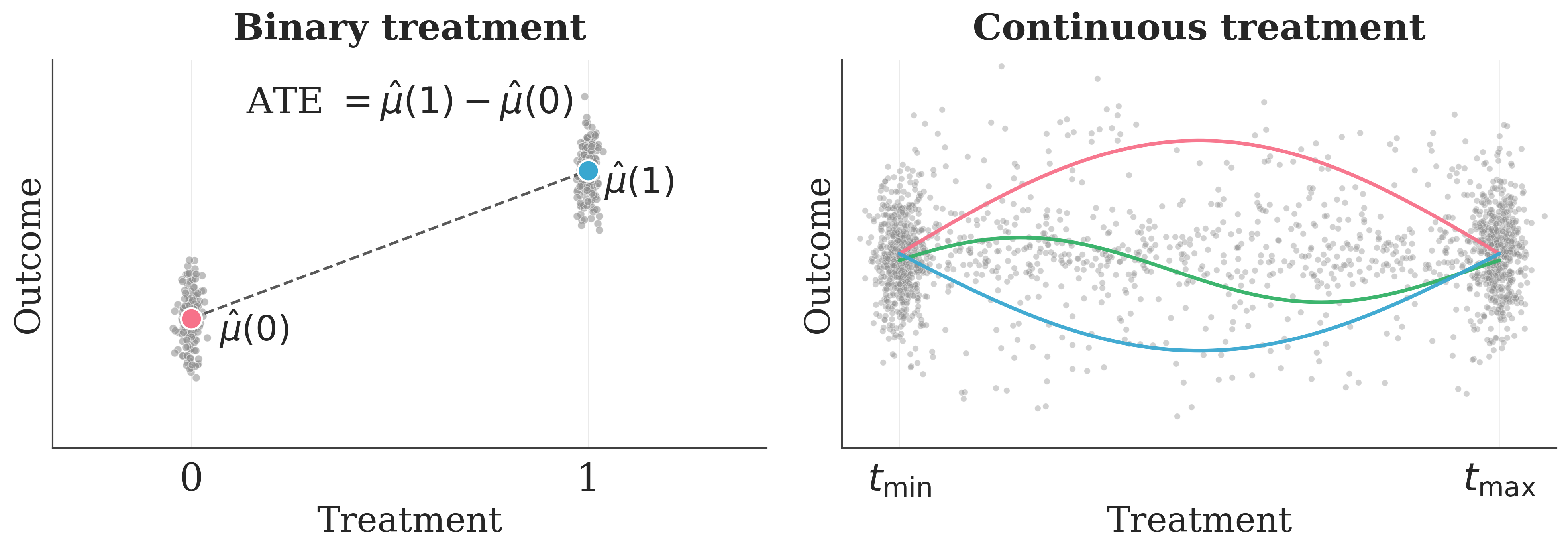}
    \caption{Estimating causal effects for continuous treatments (right) is much more challenging than for binary treatments (left), as multiple treatment-response curves fit the observed data equally well.
    }
    \label{fig:identifiability}
    \vspace{-10pt}
\end{figure*}

\section{Introduction}
\label{sec:intro}
\vspace{-6pt}
Causal inference is a central task for decision-making across many domains, including precision medicine \cite{alaa2017, pmlr-v70-shalit17a}, econometric policy-making \cite{athey2017, chernozhukov2017doubledebiased}, and algorithmic marketing \cite{bottou2013, gordon2019comparison}. Estimating the effect of an intervention from observational data alone is complicated, as the presence of confounders can bias naive estimators of potential outcomes. The causal inference community has built a rich library of estimators under the framework of ignorability, which assumes no unobserved confounding exists \cite{neal2020causal}. However, the end-to-end implementation of these estimators involves considerable time and effort: for any given task, a domain expert must inspect the data, propose an underlying mechanism to model it, choose an estimator that fits this mechanism, and only then train their model.

In addition to this approach, there has been recent work at the intersection of causal inference and meta-learning \cite{bynum2025black}. The goal here is to train a model to perform a wide variety of causal inference tasks. A particularly promising framework has been Bayesian inference and in-context learning (ICL) \cite{do-pfn, causalpfn, causalfm}. Here, a model is trained over a diverse set of causal data-generating processes (DGPs) drawn from a prior $\pi$ on possible DGPs, learning how to approximate the posterior-predictive distribution (PPD) for any new dataset. At inference, observational data for a given \emph{unseen} task is passed to the model as context, from which the model learns the PPD for the causal estimand of interest, amortizing the cost of posterior inference. This pipeline turns the typically expensive and manual approach to causal inference into a completely data-driven process.

Most previous work in Bayesian causal inference focuses on the binary treatment setting, where units can be split into control and treatment groups. However, numerous applications deal with the \emph{continuous treatment} setting, in which interventional variables have a continuous range. In marketing and economics, one is interested in the sensitivity of economic indicators to prices or rates; in pharmacology, dose-response curves are frequently used to characterize the influence of medication; and in medical contexts, one is interested not just in whether or not to treat a patient, but \emph{how much} of a medication to give.

Transitioning from binary to continuous treatments introduces significant theoretical and practical hurdles. Whereas in the binary case the goal is to estimate a single number per unit or per dataset (such as the Average Treatment Effect), continuous settings often require the estimation of a full \textit{treatment-response curve}. This is a non-trivial task, as illustrated in Figure \ref{fig:identifiability}.  Many curves are consistent with the observed data, and without additional structural assumptions the treatment-response function is not identified from observational data alone. Designing a prior that captures a suitably diverse set of continuous-valued DGPs is also highly non-trivial. The model must not only learn how to represent scalar effects, but also a wide spectrum of functions while remaining robust to the issues arising from the higher dimensionality of continuous treatment spaces. Finally, within a transformer-based architecture, representing continuous-valued treatments requires careful consideration. Standard tokenization or simple linear projections may fail to capture the high-frequency variations or local smoothness required for accurate causal discovery. 

We introduce CCPFN (\textbf{C}ontinuous \textbf{C}ausal \textbf{P}rior-\textbf{F}itted \textbf{N}etwork), a causal foundation model that directly learns to reconstruct  continuous individual treatment-response curves via in-context learning, and that can be applied off-the-shelf to any causal inference problem without fine-tuning. Central to this work is the construction of a novel prior over DGPs which interweaves three neural networks to generate covariates features, treatments, and outcomes. This prior directly encodes the assumption of identifiability via strong ignorability.

To summarize, our key contributions are: 
\begin{itemize}
[noitemsep,topsep=0pt,leftmargin=*]
    \item The first causal foundation model for continuous treatments, achieving top performance on treatment-response curve reconstruction tasks.
    \item A novel prior over data-generating processes which generates a rich set of synthetic causal data with continuous treatments.
    \item A (semi-)synthetic causal scenario generation method to facilitate model validation in causal inference with continuous treatments.
\end{itemize}

\section{Background}
\label{sec:background}
We work in the potential outcomes framework of Neyman-Rubin \cite{rubin1984bayesianly} in the \emph{continuous treatment} regime. We let capital letters denote random variables, while lower case letters denote their realizations. Let $\mathcal{T} \subseteq \mathbb{R}$ be an interval; this represents the continuous treatment space. We let $T \in \mathcal{T}$ denote the observed treatment and $\mathbf{X} \in \mathcal{X}$ the observed covariates. For each $t \in \mathcal{T}$, we denote the potential outcome under treatment $t$ by $Y^t$. We let $Y \in \mathcal{Y}$ denote the observed (factual) outcome, so that $Y = Y^T$. Finally, we let $P = P(\mathbf{X}, T, \{Y^t\}_{t \in \mathcal{T}}, Y)$ denote the \emph{data-generating process} (DGP), which is the \emph{interventional distribution}. Its marginal $P_{\text{obs}} = P(X, T, Y)$ is the \emph{observed distribution}.

Suppose we have an i.i.d.\ sample of units indexed by $n = 1, 2, \ldots, N$. For each unit $n$ we observe the covariate vector $\mathbf{x}_n$, the applied treatment $t_n$, and the observed outcome $y_n$. We are primarily concerned with recovering the \emph{individual treatment-response curve} (ITRC) which represents the causal effect of applying any of the continuous treatment levels on a unit with covariates $\mathbf{x}$. We define this curve using \emph{conditional expected potential outcomes} (CEPOs). For a given covariate vector $\mathbf{x}$ and a treatment level $t$, the CEPO is defined as 
\begin{equation}\label{eq:def_cepo}
    \mu_t(\mathbf{x}) \coloneqq \mathbb{E}[Y^t \mid \mathbf{X} = \mathbf{x}],
    \qquad \forall\, t \in \mathcal{T}.
\end{equation}
The ITRC for an individual with covariates $\mathbf{x}$ is simply the function $\mathcal{T} \to \mathcal{Y}$ traced out by the CEPO across all treatment levels: 
\begin{equation}
    t \mapsto \mu_t(\mathbf{x}), \qquad \forall\, t \in \mathcal{T}.
\end{equation}
The CEPO, and hence the ITRC, is \textit{identifiable} when it can be written as a function of the observational distribution $P_\text{obs}$ \cite{neal2020causal}. We make the following assumptions to ensure this is the case:
\begin{enumerate}[noitemsep,topsep=0pt,leftmargin=*]
    \item \textbf{Unconfoundedness}. \textit{Conditional on the covariates 
$\mathbf{X}$, the potential outcomes $Y^t$ are independent of the treatment 
assignment $T$,}
\[
    Y^t \perp T \mid \mathbf{X} \quad \forall\, t \in \mathcal{T}.
\]
    \item \textbf{Positivity/Overlap}. \textit{There exists a constant $c > 0$ such 
that for all $\mathbf{x} \in \mathcal{X}$ and all $t \in \mathcal{T}$,}
\[
    p_{T \mid \mathbf{X}}(t \mid \mathbf{x}) \geq c.
\]
\end{enumerate}
Under Assumptions 1--2, the CEPO (and hence the ITRC) is identifiable from the observational 
distribution \cite{hirano2004propensity, neal2020causal}:
\[
    \mu_t(\mathbf{x}) = \mathbb{E}[Y^t \mid \mathbf{X} = \mathbf{x}] 
    = \mathbb{E}[Y \mid \mathbf{X} = \mathbf{x},\, T = t].
\]

\textbf{Bayesian Causal Inference.} A Bayesian formulation of causal inference places a prior $\pi(\psi)$ over DGPs $\psi$, each indexing a joint distribution $P^\psi(\mathbf{X}, T, \{Y^t\}_{t \in \mathcal{T}}, Y)$. Given i.i.d.\ observations $\mathcal{D}_\text{obs} = \{(\mathbf{x}^{(n)}, t^{(n)}, y^{(n)})\}_{n=1}^N$ sampled from the observational distribution $P^\psi_\text{obs}$, Bayes' rule yields the posterior distribution $\pi(\psi \mid \mathcal{D}_\text{obs})$. The posterior-predictive distribution (PPD) of any causal estimand 
$g(\psi)$ is then defined as
\begin{equation}
    \pi^g(\cdot \mid \mathcal{D}_{\text{obs}}) := \left[ B \mapsto \int \mathbb{I}(g(\psi) \in B) \, \pi(\psi \mid \mathcal{D}_{\text{obs}}) \, \mathrm{d}\psi \right], \qquad B \in \mathcal{B}, \label{eq:def_ppd}
\end{equation}
where $\mathcal{B}$ is the Borel $\sigma$-algebra over $\mathbb{R}$. This paradigm provides a unified framework for point estimation ~\cite{rubin1984bayesianly, oganisian2021practical, li2023bayesian}. In practice, computing the posterior is intractable, motivating the amortized approach to estimating the PPD described below.

\textbf{Prior-Fitted Networks.} Recent works \cite{hollmann2023tabpfn, ma2025tabdpt, tabicl, causalpfn, do-pfn, causalfm} on PFNs amortize posterior-predictive inference using a single transformer-based \cite{vaswani2017attention} neural network $q_\theta$, trained to directly approximate the PPD via the data-prior loss
\begin{equation}
    \ell(\theta) := \mathbb{E}_{\varphi \sim \pi,\, \mathcal{D} \cup \{x,y\} 
    \sim P^\varphi}\bigl[-\log q_\theta(y \mid x, \mathcal{D})\bigr]. \label{eq:data-prior-loss_def}
\end{equation}
Training requires only prior samples, while no posterior sampling is needed at inference time.

As in \cite{causalpfn} our target is the PPD of the CEPO (i.e. $g(\psi) = \mu_t(\mathbf{x}; \psi)$ in \Cref{eq:def_ppd})
where our CEPO is parameterised by the DGP index $\psi$. Balazadeh et al.~\cite{causalpfn} proved that 

\[\mathbb{E}_{\mu \sim \pi^{\mu_t}(\cdot \mid \mathbf{x}, \mathcal{D}_{\text{obs}})}[\mu] 
\xrightarrow{a.s.} \mu_t(\mathbf{x} \,;\, \psi^\star), 
\qquad \forall t \in \mathcal{T}, \text{ and almost all } \mathbf{x} \in \mathcal{X},\] 

provided the prior $\pi$ is CEPO-identifiable. Following their framework, we train a transformer $q_\theta$ to approximate the full predictive distribution $\pi^{\mu_t}$ using a \emph{Causal Data-Prior Loss}. For any $t \in \mathcal{T}$, this loss is defined as
\begin{equation}
    \mathcal{L}_t(\theta) := \mathbb{E}_{\psi \sim \pi,\, \mathcal{D}_{\text{obs}} \cup \{\mathbf{x}\} \sim P_{\text{obs}}^{\psi}} \left[ -\log q_\theta(\mu_t(\mathbf{x} \,;\, \psi) \mid \mathbf{x}, t, \mathcal{D}_{\text{obs}}) \right]. \label{eq:causal-loss_def}
\end{equation}
During training, we randomly sample $t \sim \text{Uniform}(\mathcal{T})$ in order to cover the full treatment range $\mathcal{T}$. Moreover, in practice, we use regression-as-classification and quantize the outcome variable into $L = 1024$ bins from $[-10, 10]$ (after $z$-standardization). The model then outputs $q_\theta$ as a discrete histogram distribution on each of these $L$ bins. To approximate the true CEPO-PPD in a tractable manner we assume it is given by $\mathcal{N}(\mu_t(\mathbf{x}), \sigma^2)$ for $\sigma \ll 1$. The \emph{histogram loss} which we train with is then defined as
\begin{equation}
    \mathcal{L}_{\text{HL}} (\theta \mid \mathbf{x}, t) = -\sum_{l = 1}^L \mathcal{N}(\mu_t(\mathbf{x}), \sigma^2)[l] \log q_\theta(l \mid \mathbf{x}, t, \mathcal{D}_{\text{obs}}),  \label{eq:histogram-loss}
\end{equation}
where $\mathcal{N}(\mu_t(\mathbf{x}), \sigma^2)[l]$ denotes the probability of a $\mathcal{N}(\mu_t(\mathbf{x}), \sigma^2)$-random variable lying in bin $l$, and $q_\theta(l \mid \mathbf{x}, t, \mathcal{D}_{\text{obs}})$ denotes the model's output probability for bin $l$.

\textbf{Related Work.} While ours is the first model to leverage in-context learning for the continuous treatment regime in causal inference, numerous techniques have been developed over the past decade to address this regime due to its importance across a variety of domains. DRNet \cite{Schwab_2020} addresses the challenge of multi-armed treatments, each with associated treatment parameters. For each treatment, DRNet breaks the treatment range into $E \in \mathbb{N}$ equally-sized subintervals and assigns a head to each. This ensures the effect of the treatment parameter persists throughout the network layers, but it can lead to discontinuities in the learned treatment-response curves. SCIGAN \cite{scigan_2020} leverages generative adversarial networks \cite{goodfellow2014gan} to generate and learn counterfactual representations. VCNet \cite{vcnet_2021} further addresses the continuous treatment space by training a varying coefficient  network, where neural network parameters $\theta = \theta(t)$ depend on the treatment value $t$. ADMIT \cite{admit_2022} extends the theoretical framework by bounding the counterfactual loss of estimating treatment-response curves and proposes an algorithm that makes counterfactual estimations. Meanwhile, several groups have approached the binary treatment setting for causal inference via causal foundation modeling; see \cite{do-pfn, causalpfn, causalfm}. 

Training on synthetic data is a common practice in the tabular foundation model (TFM) space. In particular, TabPFN \cite{hollmann2023tabpfn, tabpfn2.5} and TabICL \cite{tabicl, tabiclv2} both train on purely synthetic data generated from a carefully-designed prior. While not designed specifically for causal inference, their priors (like ours) are generated via synthetic structural causal models (SCMs) using randomly-generated neural networks. More recently, it has been shown that data augmentation during fine-tuning with synthetic, SCM-based causal data can improve TFM performance \cite{buhlercausalaugmentation}.

\vspace{-4pt}
\section{Method}
\label{sec:method}
\vspace{-4pt}
\textbf{Generating a Prior.} Following the general approach of CausalPFN~\cite{causalpfn}, we design a novel prior over potential DGPs arising from continuous treatment scenarios. As we are interested in the setting where the causal estimands are \emph{identifiable}, we design these priors so that the assumptions of unconfoundedness and positivity hold (see Section \ref{sec:background}). The other key requirement of a training prior is to generate a suitably \emph{diverse} set of DGPs so as to represent a variety of real-world scenarios. This is complicated considerably when dealing with continuous treatments $\mathcal{T}$, and so the design of our prior differs considerably from \cite{causalpfn}. 

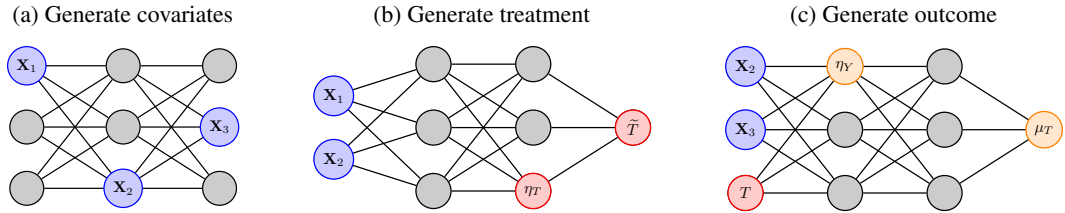
\begin{figure}[ht]
    \centering
\begin{subfigure}[t]{0.22\textwidth}
\centering
\caption{Generate covariates } 
\vspace{5pt}
\resizebox{\textwidth}{!}{%
  \begin{tikzpicture}[
    x=2.2cm, y=1.4cm,
    blacknode/.style={thick, draw=black, fill=black!20, circle, minimum size=22},
    bluenode/.style={thick, draw=blue, fill=blue!20, circle, minimum size=22}
  ]
    \readlist\Nnod{3, 3, 3} 
    
    \foreachitem \N \in \Nnod{ 
      \foreach \i [evaluate={\x=\Ncnt; \y=\N/2-\i+0.5; \prev=int(\Ncnt-1);}] in {1,...,\N}{ 
        \node[blacknode] (N\Ncnt-\i) at (\x,\y) {};
        
        \ifnum\Ncnt>1 
          \foreach \j in {1,...,\Nnod[\prev]}{ 
            \draw[thick] (N\prev-\j) -- (N\Ncnt-\i); 
          }
        \fi 
      }
    }
    
    \node[bluenode] at (N1-1) {$\mathbf{X}_{1}$};
    \node[bluenode] at (N2-3) {$\mathbf{X}_{2}$};
    \node[bluenode] at (N3-2) {$\mathbf{X}_{3}$};
    
  \end{tikzpicture}%
}
\label{fig:step1}
\end{subfigure}%
\hfill
\begin{subfigure}[t]{0.32\textwidth}
\centering
\caption{Generate treatment} 
\vspace{5pt}
\resizebox{\textwidth}{!}{%
  \begin{tikzpicture}[
    x=2.2cm, y=1.4cm,
    blacknode/.style={thick, draw=black, fill=black!20, circle, minimum size=22},
    orangenode/.style={thick, draw=orange, fill=orange!20, circle, minimum size=22},
    bluenode/.style={thick, draw=blue, fill=blue!20, circle, minimum size=22},
    rednode/.style={thick, draw=red, fill=red!20, circle, minimum size=22}
  ]
    \readlist\Nnod{2, 3, 3, 1} 
    
    \foreachitem \N \in \Nnod{ 
      \foreach \i [evaluate={\x=\Ncnt; \y=\N/2-\i+0.5; \prev=int(\Ncnt-1);}] in {1,...,\N}{ 
        \node[blacknode] (N\Ncnt-\i) at (\x,\y) {};
        
        \ifnum\Ncnt>1 
          \foreach \j in {1,...,\Nnod[\prev]}{ 
            \draw[thick] (N\prev-\j) -- (N\Ncnt-\i); 
          }
        \fi 
      }
    }
    
    \node[bluenode] at (N1-1) {$\mathbf{X}_{1}$};
    \node[bluenode] at (N1-2) {$\mathbf{X}_{2}$};

    \node[rednode] at (N4-1) {{$\widetilde T$}};
    \node[rednode] at (N3-3) {{$\eta_T$}};
    
  \end{tikzpicture}%
}
\label{fig:step2}
\end{subfigure}%
\hfill
\begin{subfigure}[t]{0.32\textwidth}
\centering
\caption{Generate outcome} 
\vspace{5pt}
\resizebox{\textwidth}{!}{%
  \begin{tikzpicture}[
    x=2.2cm, y=1.4cm,
    blacknode/.style={thick, draw=black, fill=black!20, circle, minimum size=22},
    orangenode/.style={thick, draw=orange, fill=orange!20, circle, minimum size=22},
    bluenode/.style={thick, draw=blue, fill=blue!20, circle, minimum size=22},
    rednode/.style={thick, draw=red, fill=red!20, circle, minimum size=22}
  ]
    \readlist\Nnod{3, 3, 3, 1} 
    
    \foreachitem \N \in \Nnod{ 
      \foreach \i [evaluate={\x=\Ncnt; \y=\N/2-\i+0.5; \prev=int(\Ncnt-1);}] in {1,...,\N}{ 
        \node[blacknode] (N\Ncnt-\i) at (\x,\y) {};
        
        \ifnum\Ncnt>1 
          \foreach \j in {1,...,\Nnod[\prev]}{ 
            \draw[thick] (N\prev-\j) -- (N\Ncnt-\i); 
          }
        \fi 
      }
    }
    
    \node[bluenode] at (N1-1) {$\mathbf{X}_{2}$};
    \node[bluenode] at (N1-2) {$\mathbf{X}_{3}$};
    \node[rednode] at (N1-3) {$T$};

    \node[orangenode] at (N4-1) {{$\mu_T$}};
    \node[orangenode] at (N2-1) {{$\eta_Y$}};
    
  \end{tikzpicture}%
}
\label{fig:step3}
\end{subfigure}

\vspace{0.5cm} 
\label{fig:all_steps}
    \vspace{-15pt}
    \caption{A schematic of our 3-MLP prior. In practice all MLPs drop edges with a certain probability.}
    \label{fig:our-prior}
\end{figure}
Inspired by the construction of structural causal models (SCMs) via randomized MLPs in \cite{tabiclv2}, our prior consists of three separate randomized MLPs (see Figure \ref{fig:our-prior}):
\begin{itemize}[noitemsep,topsep=0pt,leftmargin=*]
    \item $\mlpx$ generates the covariates $\mathbf{X}$, which can be divided into three disjoint subsets: the set $\mathbf{X}^{T}$ which are direct causes of $T$ only; the set $\mathbf{X}^Y$ which are direct causes of $Y$ only; and the set of confounders $\mathbf{X}^{\text{conf}}$ which are direct causes of both $T$ and $Y$.
    \item $\mlpt$ generates the observed treatment $T$ from $\mathbf{X}^T \cup \mathbf{X}^{\text{conf}}$.
    \item $\mlpy$ generates outcomes (both factual and counterfactual), given treatment $t \in \mathcal{T}$ and covariates $\mathbf{X}^Y \cup \mathbf{X}^{\text{conf}}$.
\end{itemize}
Each SCM produces samples 
\begin{align*}
    \big\{ \big( \mathbf{x}_n, t_n, y_n, t_n', \mu_{t_n'}(\mathbf{x}_n) \big) \big\}_{n = 1}^N
\end{align*}
where $(t_n, y_n)$ are the factual (observed) treatment and outcome, respectively, $t_n'$ is a counterfactual treatment, and $\mu_{t_n'}(\mathbf{x}_n)$ is the corresponding CEPO. We let $N$ denote the number of samples throughout.

We presently discuss each component MLP; further details can be found in Appendix~\ref{sec:app-b}. The following hyperparameters are randomly sampled: 
\begin{itemize}[noitemsep,topsep=0pt,leftmargin=*]
    \item The number of layers $L_{\mathbf{X}}, L_T$, and $L_Y$ of each MLP;
    \item The number of hidden units $H_{\mathbf{X}}, H_T$, and $H_Y$ of each MLP;
    \item The densities $d_{\mathbf{X}}, d_T$, and $d_Y$ are uniformly sampled in $[0.1, 1]$; edges in each $\textbf{MLP}_{(\cdot)}$ are dropped with probability $1 - d_{(\cdot)}$;
    \item The degree of confounding $\rho \sim \mathcal{U}(0, 1)$, which is the proportion of covariates $\mathbf{X}$ that are direct causes of both $T$ and $Y$. The remaining covariates are randomly split into $\textbf{X}^T$ and $\textbf{X}^Y$.
\end{itemize}

$\mlpx$\textbf{.} In addition to $L_{\mathbf{X}}, H_{\mathbf{X}}$, and $d_{\mathbf{X}}$, a noise scale $s > 0$ and a natural number $K > 0$ (representing the number of covariates) are chosen. For each $l \in [L_{\mathbf{X}}]$, noise values $\eps^{(l)} \in \R^N \times \R^{H_{\mathbf{X}}}$ for each $n \in [N]$ are generated by randomly-chosen distributions in $\{\mathcal{N}(0, 1), \text{Laplace}(0, 1), t_3\}$ followed by a random shift and scaling. The input $\eps^{(0)}$ is propagated through the MLP by randomly initialized weights:
\begin{equation}
    z^{(l)} = {\sigma}^{(l)}(W^{(l)}z^{(l-1)}) + \epsilon^{(l)}, \quad z^{(0)} = \epsilon^{(0)},  \label{eq:mlp_x}
\end{equation}
where each $\sigma^{(l)}$ represents a vector of random activation functions. The MLP weights come from the normal distribution $\mathcal{N}(0, \sigma_w^2)$, where $\sigma_w = \big(\frac{2}{\text{max}(H * p_d, 1.0)}\big)^\frac{1}{2}$; $H$ is the hidden width of the MLP while $p_d$ represents the probability of a given weight being kept by a random sparsity mask. We then choose $K$ nodes among the $z^{(l)}$ to be covariates; this produces covariates differing in complexity and marginal distribution. In addition, to simulate realistic tabular data, we apply random ``tabular corruption''. A random subset of covariate nodes is either binarized, quantized, or zero-inflated (across the $N$ samples). Contrary to past works \cite{hollmann2023tabpfn, tabiclv2}, which only apply this transformation \emph{after} the forward pass, we transform roughly 35\% of covariate nodes \emph{during} the forward pass. This allows the SCM to see and utilize the realistic tabular features during generation, adding realistic tabular diversity to the DGP itself, rather than solely post-hoc. Another roughly 65\% of the remaining covariates are further corrupted after the forward pass is completed. The superiority of this method as opposed to solely post-hoc tabular corruption, is demonstrated in our ablations (see Table \ref{tab:ablation-noinpass}). This does not affect the theoretical validity of the SCM, as all tabular corruption occurs within $\mlpx$ alone; it merely enhances the diversity and realism of the prior.

$\mlpt$\textbf{.} The input to $\mlpt$ is $\mathbf{X}^T \cup \mathbf{X}^{\text{conf}}$, and the output is a scalar $\tilde T$ which is the expected treatment value given $\mathbf{X}^T \cup \mathbf{X}^{\text{conf}}$. A random node $\eta_T$ in the hidden layers is chosen to act as a heteroscedastic noise scale parameter. To produce the final outcome $T$ we compute
\begin{equation}
    T = \tilde T + \sigma(\tilde T) \cdot \eta_T \cdot \eps, \qquad \eps \sim \mathcal{N}(0, 1),
\end{equation}
where $\sigma(\tilde T)$ is the empirical standard deviation of the $\tilde T$ over the $N$ samples. Rather than designating the output node as $T$ itself, this method practically ensures that positivity $p(T \mid \mathbf{X} = \bfx) > 0$ holds, as $T$ is centered at $\tilde T$ with strictly positive variance. Finally, the treatment $T$ is min-max scaled to lie in $[0, 1]$. 

$\mlpy$\textbf{.} The input to $\mlpy$ is $\mathbf{X}^Y \cup \mathbf{X}^{\text{conf}}$ as well as a treatment value $t \in [0, 1]$. This MLP can thus generate both factual and counterfactual outcomes. Similar to $\mlpt$, the output is a scalar which is the CEPO $\mu_t(\mathbf{X})$, while a random node $\eta_Y$ in the hidden layers is chosen as noise scale parameter. During the factual forward pass, the output $T$ of $\mlpt$ is passed as input, and the factual outcome $Y$ is generated by 
\begin{equation}
    Y = \mu_T(\textbf{X}) + \sigma(\mu) \cdot \eta_Y \cdot \eps, \qquad \eps \sim \mathcal{N}(0, 1),
\end{equation}
where $\sigma(\mu)$ is the empirical standard deviation of the CEPOs over the $N$ samples. To generate CEPO for counterfactual treatment values $t \neq T$, one passes $t$ as input and selects the output node $\mu_t(\mathbf{X})$ without adding the noise $\eta_Y$. 

\begin{wrapfigure}[9]{r}{0.3\textwidth}
    \centering
    \vspace{-0pt}
\begin{tikzpicture}
  \tikzset{
    graphnode/.style={circle, draw, thick, minimum size=1.2cm, inner sep=0pt},
    bluenode/.style={circle, fill=blue!20, draw=blue},
    rednode/.style={circle, fill=red!20, draw=red},
    orangenode/.style={circle, fill=orange!20, draw=orange}
  }
  
  \node[rednode] (T) at (0,0) {$T$};
  \node[bluenode] (X) at (1,5/3) {$X$};
  \node[orangenode] (Y) at (2,0) {$Y$};
  
  \draw[->, >=stealth, thick] (X) -- (T);
  \draw[->, >=stealth, thick] (X) -- (Y);
  \draw[->, >=stealth, thick] (T) -- (Y);
  
\end{tikzpicture}       
    \caption{Causal graph associated with the backdoor setting.}
    \label{fig:causal-graph-1}
\end{wrapfigure}
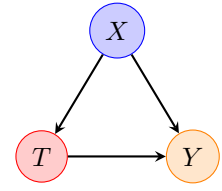
In summary, this method constructs a prior over possible DGPs which arise in the \emph{backdoor setting} of causal inference, with causal graph as shown in Figure \ref{fig:causal-graph-1}.

We also designed two additional priors that employed polynomial basis sampling and value-based sampling from MLP-generated tables. These priors are further described in Appendix~\ref{app:alt-prior}, and we include them in ablation studies. 

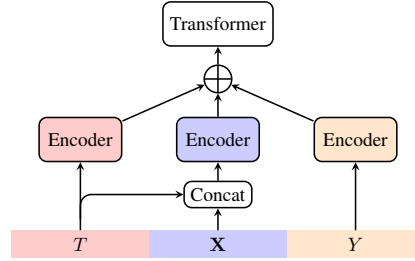
\begin{wrapfigure}[14]{r}{0.4\textwidth}
    \centering
    \vspace{-6pt}
\resizebox{\linewidth}{!}{%
  
  

  



    
\begin{tikzpicture}[>=stealth, thick]

  \fill[red!20] (-1.25, .75) rectangle (1.25, 1.25);
  \fill[blue!20] (1.25,.75) rectangle (3.75, 1.25);
  \fill[orange!20] (3.75, .75) rectangle (6.25, 1.25);

  \node (T) at (0,1) {$T$};
  \node (X) at (2.5,1) {$\mathbf{X}$};
  \node (Y) at (5,1) {$Y$};

  \node[draw, rectangle, rounded corners] (Concat) at (2.5,1.9) {Concat};

  \node[
    draw, rectangle, rounded corners,
    minimum width=1.5cm,
    minimum height=8mm,
    fill=red!20
  ] (Et) at (0,2.9) {Encoder};

  \node[
    draw, rectangle, rounded corners,
    minimum width=1.5cm,
    minimum height=8mm,
    fill=blue!20
  ] (Ex) at (2.5,2.9) {Encoder};

  \node[
    draw, rectangle, rounded corners,
    minimum width=1.5cm,
    minimum height=8mm,
    fill=orange!20
  ] (Ey) at (5,2.9) {Encoder};

  \draw[->] (X) -- (Concat);
  \draw[->] (Y) -- (Ey);
  \draw[->] (Concat) -- (Ex);

  \draw (T) -- (0,1.45) coordinate (split);
  \draw[->] (split) -- (Et);
  \draw[->, rounded corners=2mm] (split) |- (Concat);

  \node[sum] (sum) at (2.5,4.0) {};

  \node[
    draw,
    rectangle,
    rounded corners,
    minimum width=2cm,
    minimum height=8mm
  ] (Transformer) at (2.5,5.0) {Transformer};

  \draw[->] (Et) -- (sum);
  \draw[->] (Ex) -- (sum);
  \draw[->] (Ey) -- (sum);
  \draw[->] (sum) -- (Transformer);

\end{tikzpicture}
}
\caption{Illustration of the tri-encoder schematic used by CCPFN. Treatments $T$ are additionally routed through a separate encoder to boost treatment signal.}
    \label{fig:model-arch}
\end{wrapfigure}

\textbf{Model Architecture.} We use a similar architecture to CausalPFN~\cite{causalpfn}, with a PFN-style transformer encoder that leverages in-context learning to learn the parameterized CEPO-PPD $q_\theta$. The central new architectural piece of our method CCPFN is a separate encoder for treatments to ensure that the signal from the treatment variable is not lost in high-dimensional settings. Input tokens $t$ are passed directly through a nonlinear $T$-encoder, and in addition are separately appended to $\bfx$ and passed through a linear encoder (reminiscent of S-Learners~\cite{slearner}); see Figure \ref{fig:model-arch}. Moreover, as we work in the continuous treatment setting, we $z$-standardize all outcomes together (rather than splitting into control/treatment groups) as well as all treatment values. Given a query unit $\mathbf{x}$, a grid of treatment values can be passed to CCPFN as a batch to estimate the full range of the ITRC $t \mapsto \mu_t(\bfx)$ in a single forward pass.

\textbf{Training.} At each step of training, a DGP $\psi \sim \pi$ is sampled to yield a SCM which is generated by the three MLPs described above. We generate a dataset
\begin{align*}
    \big\{ \big( \mathbf{x}_n, t_n, y_n, t_n', \mu_{t_n'}(\mathbf{x}_n) \big) \big\}_{n = 1}^N
\end{align*}
of both factual and counterfactual scenarios. Counterfactual treatments are sampled uniformly on $[0, 1]$. This ensures that the loss in \eqref{eq:causal-loss_def} is minimized over the whole treatment range $\mathcal{T} = [0, 1]$. At training time, the dataset is randomly shuffled and a context/query split position $M$ is chosen. The observational dataset $\mathcal{D}_{\text{obs}} = \{(\bfx_n, t_n, y_n)\}_{n = 1}^M$ is passed as context, and the counterfactuals $\{(\bfx_n, t_n')\}_{n = {M + 1}}^N$ are passed as the query.

\section{Experimental Setup}
\label{sec:experiments}
\vspace{-4pt}
The fundamental problem of causal inference---namely, the inability to observe counterfactual outcomes---poses a unique challenge to experiments and model evaluation in causal inference \cite{holland_1986}. In the absence of randomized controlled trials (RCTs), which are often impractical or unethical in various domains, causal inference models are tested on synthetic or semi-synthetic data \cite{kallus_zhou_2018, realcause, admit_2022, causal-pipeline-2025}. We follow this approach, implementing several synthetic and semi-synthetic benchmarks for both our validation and test datasets. 

\textbf{Scenario-Generation Pipeline.} While in the binary treatment regime public benchmarks have become more available recently, there is a serious lack of a broad suite of benchmarks in the continuous treatment regime. Part of the difficulty lies in having to store a potentially infinite number of counterfactuals per individual. To address these issues, we created a (semi-)synthetic scenario-generation pipeline. Each scenario represents a causal inference scenario/context (for example, the administration of a particular drug to a particular set of patients). The base covariates $\mathcal{D}_{\text{cov}} = \{\bfx_n\}_{n = 1}^N$ are loaded in from either an existing synthetic generation mechanism or from real-world tabular data. We then construct propensity and outcome models for each scenario based on the context and covariates provided. For the validation datasets, we tasked an LLM agent with generating plausible propensity models and outcome models for the given covariates (see Appendix~\ref{sec:validation-data-construction}). For the test datasets, we used or modified propensity and outcome models from previous benchmarks. This pipeline generates causal DGPs for the backdoor scenario (\Cref{fig:causal-graph-1}). We ensured that approximately 50\% of covariates were direct causes of both treatment and outcome to ensure a high degree of confounding. This confounding present in the DGPs ensures that validating against these scenarios is a genuinely causal task (rather than e.g. simply correlation matching). To verify the LLM-produced scenarios, we manually inspected each scenario and sampled the ITRCs they produced, using domain expertise to check consistency and provide a sanity check.

With this method, we generated 14 scenarios (8 validation, 6 test). These span a wide range of contexts, including drug dosage effects, algorithmic marketing, debt forgiveness strategies, and news article engagement. We include plots of example treatment-response functions from several scenarios in Figure \ref{fig:ITRC-plots}. More details can be found in Appendices \ref{sec:scenarios} and \ref{sec:validation-data-construction}.

\begin{figure}[t]
    \centering
    
    \begin{subfigure}[b]{0.4\linewidth}
        \centering
        \includegraphics[width=\linewidth]{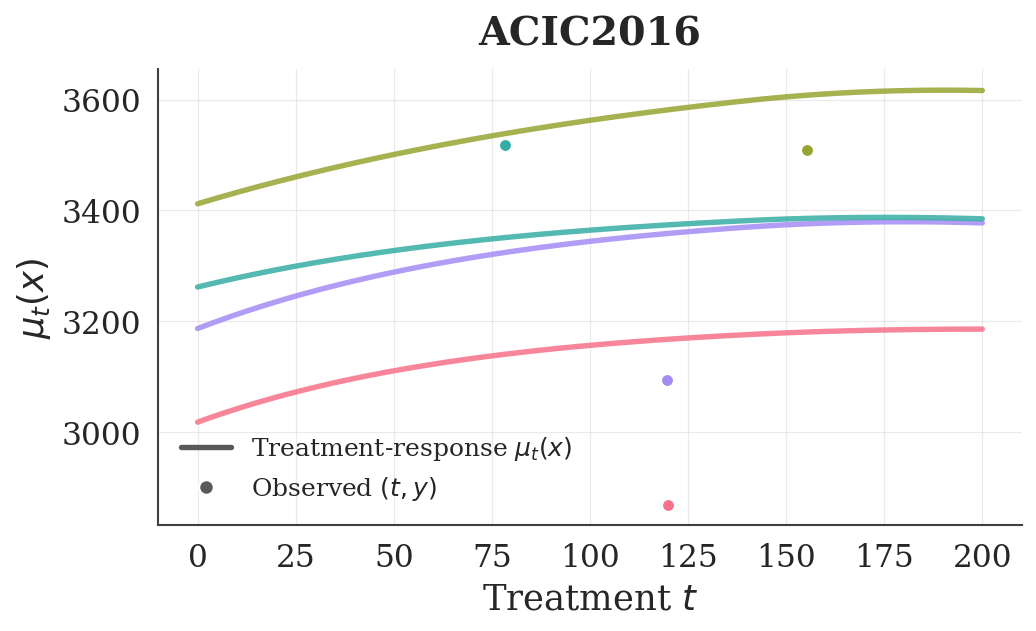}
        \label{fig:acic2016}
    \end{subfigure}
    \hspace{0.05\linewidth}
    \begin{subfigure}[b]{0.4\linewidth}
        \centering
        \includegraphics[width=\linewidth]{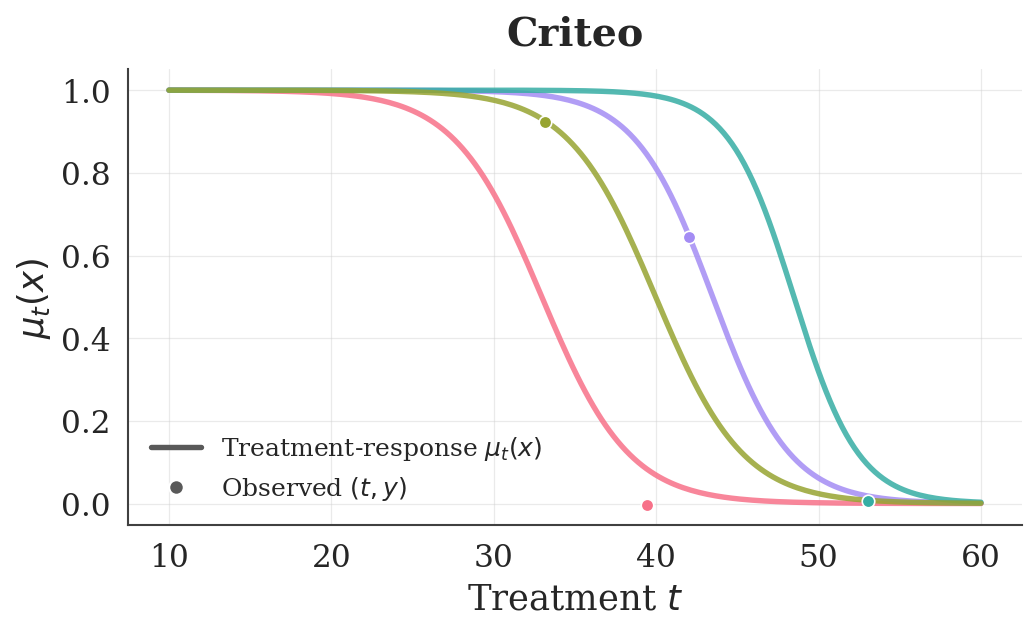}
        \label{fig:criteo}
    \end{subfigure}
    
    \begin{subfigure}[b]{0.4\linewidth}
        \centering
        \includegraphics[width=\linewidth]{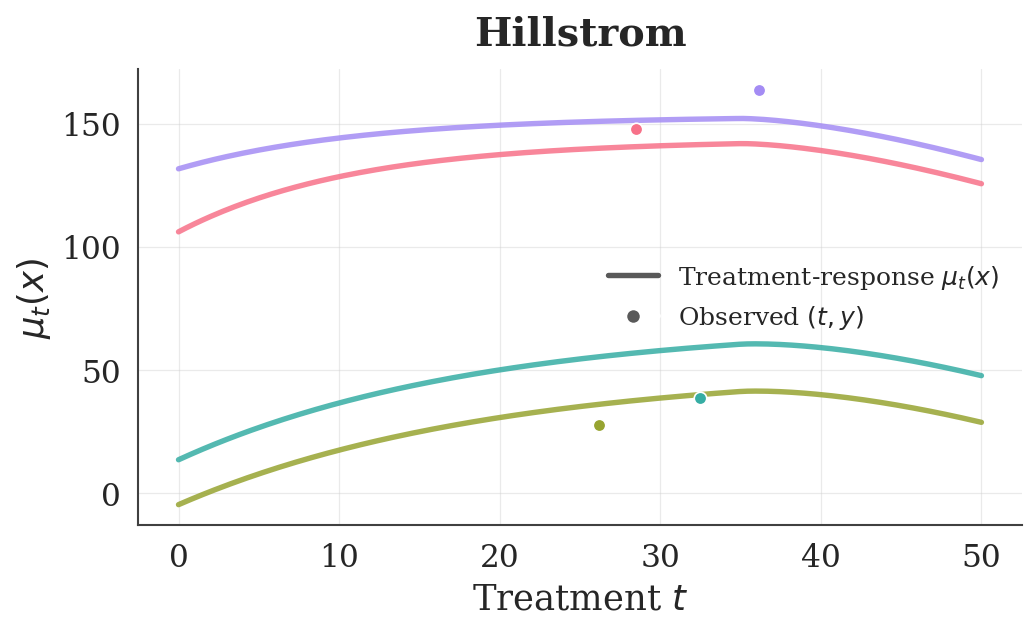}
        \label{fig:hillstrom}
    \end{subfigure}
    \hspace{0.05\linewidth}
    \begin{subfigure}[b]{0.4\linewidth}
        \centering
        \includegraphics[width=\linewidth]{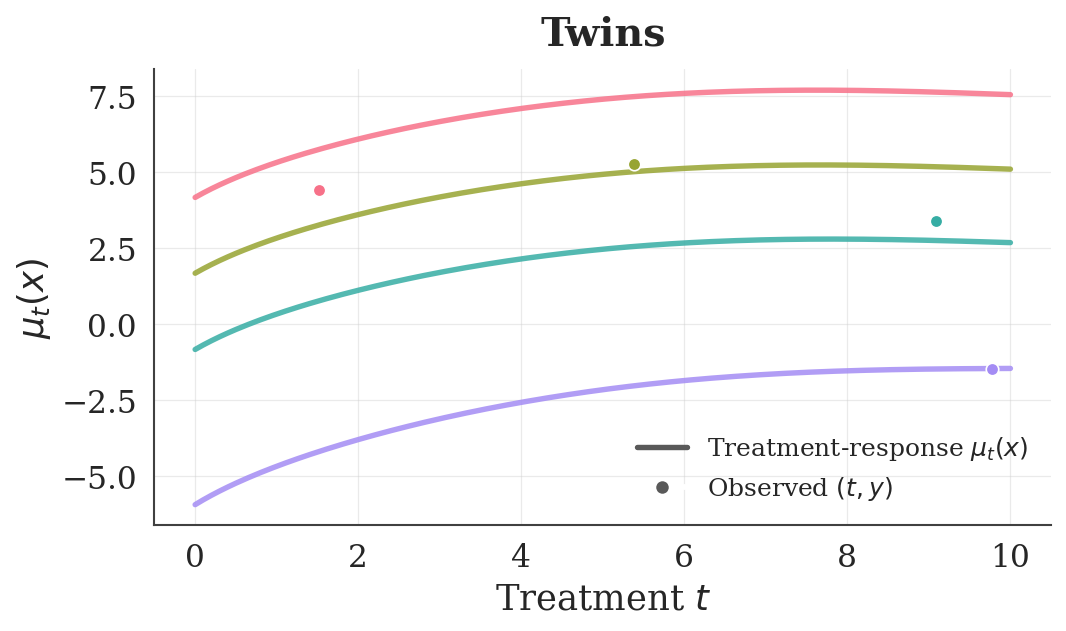}
        \label{fig:twins}
    \end{subfigure}
    \vspace{-8pt}
    \caption{Example individual treatment-response curves (ITRCs) for four of our validation scenarios (ACIC2016, Criteo, Hilltrom, and Twins). Solid curves are ITRCs for randomly-selected individuals; circles are the corresponding observed $(T, Y)$. Note that the observations do not lie precisely on the ITRCs due to the presence of exogenous noise; the curves represent expectations (CEPOs), not exact counterfactual outcomes.}
    \label{fig:ITRC-plots}
    \vspace{-16pt}
\end{figure}

\begin{wraptable}[9]{r}{7.5cm}
    \centering
        \vspace{-14pt}
    \caption{Summary of test datasets used as benchmarks.}
    \label{tab:dataset_stats}
    \setlength{\tabcolsep}{2.5pt}
\setlength{\extrarowheight}{-1pt}
\setlength{\aboverulesep}{0.5ex}
\setlength{\belowrulesep}{0.5ex}
\setlength{\cmidrulesep}{0.3ex}
        \vspace{-4pt}
    \begin{tabular}{lll}
        \toprule
        Dataset & \# Samples & \# Covariates \\ 
        \midrule
        MVICU~\cite{Schwab_2020, admit_2022}    & 4,963  & 13 \\ 
        Debt~\cite{causal-pipeline-2025}     & 10,000 & 10 \\ 
        News~\cite{Schwab_2020, admit_2022}     & 7,881  & 2,870 \\ 
        NewsHet~\cite{admit_2022}  & 7,881  & 2,870 \\ 
        TCGA~\cite{Schwab_2020, admit_2022}     & 4,428     & 4,000 \\ 
        Warfarin~\cite{IWPC2009, warfarin, kallus_zhou_2018} & 4,490  & 19 \\ 
        \bottomrule
    \end{tabular}
\end{wraptable}

\textbf{Benchmarks.} Our test data benchmarks are comprised of one fully synthetic dataset and five semi-synthetic datasets that are often used to study causal inference, and span the domains of medicine, finance, and user research. These datasets are not used for training or any model tuning. Further details about the benchmarks are provided in Appendix~\ref{sec:dataset-details}.

\noindent
\textbf{Validation Datasets.} Our validation datasets, used to tune model hyperparameters during training, were created by a mix of manual specification and LLM-generated scenarios. The prompts we used to generate DGPs can be found in Appendix \ref{sec:validation-data-construction}. We specifically make use of covariates from the ACIC2016~\cite{acic2016}, ACIC2018~\cite{acic2018}, Criteo \cite{criteo}, Hillstrom~\cite{hillstrom2008}, Lalonde~\cite{dehejia1999causal, causaldata_py}, Lenta~\cite{lenta2020bigtarget, shevchenko2020scikituplift}, Twins~\cite{realcause}, and X5~\cite{x5retailhero2019, shevchenko2020scikituplift} datasets, which are commonly employed in causal literature. 

\textbf{Metrics.} We use two primary metrics to evaluate our model's ability to perform two key tasks:
\begin{enumerate}[noitemsep,topsep=0pt]
    \item Reconstruct individual treatment-response curves;
    \item Prescribe the optimal treatment level/dosage for each individual.
\end{enumerate}
To evaluate performance on the first task, we use the mean integrated squared error (MISE) \cite{Schwab_2020} between the true treatment-response curve $\mu_t$ and the model's predicted $\hat\mu_t$, which is reported as the mean of the CEPO-PPD $q_\theta(t, \bfx)$. We normalize by the treatment interval length to standardize. This is calculated over all $N$ individuals in a given dataset and over the whole treatment range $[a, b]$:
\begin{equation}
    \text{MISE} = \frac{1}{N} \sum_{n = 1}^N \frac{1}{b - a}\int_a^b \big| \hat\mu_t(\mathbf{x}_n) - \mu_t(\mathbf{x}_n) \big|^2 \, dt,  \label{eq:mise_def}
\end{equation}
where $\mathbf{x}_n$ is the covariate vector of the $n$th individual.

To evaluate performance on the second task, we use the mean dosage policy error (DPE) \cite{Schwab_2020}. This is the error between the true  CEPO at the true optimal dosage/treatment level $t^*$ and the true CEPO at the model's estimated optimal dosage $\hat{t}^*$, averaged over the whole dataset:
\begin{equation}
    \text{DPE} = \frac{1}{N}\sum_{n = 1}^N \big(\mu_{t^*(\mathbf{x}_n)}(\mathbf{x}_n) - \mu_{\hat{t}^*(\mathbf{x}_n)}(\mathbf{x}_n)\big)^2,  \label{eq:dpe_def}
\end{equation}
where $t^*(\mathbf{x})$ is the optimal dosage for an individual with covariates $\mathbf{x}$. We consider in this paper optima that are either global minima or maxima, depending on the scenario. The estimated optimal dosage $\hat t^*(\mathbf{x})$ of a model is calculated by first reconstructing the model's estimate of the full treatment-response curve $t \mapsto \mu_t(\mathbf{x})$ on a finite mesh and choosing $\hat t^*(\mathbf{x})$ to be the argmin/argmax. 

The two metrics are clearly related, while capturing different aspects of model performance: MISE measures the model's ability to accurately reconstruct the shape of the entire treatment-response curve over the full range of possible treatments (Task 1), while DPE measure the model's ability to accurately prescribe an optimal treatment level for a given individual (Task 2).

\textbf{Baseline Methods.} We compare CCPFN to various causal inference models which are trained on each individual dataset. We use neural network-based methods, including ADMIT \cite{admit_2022}, SCIGAN \cite{scigan_2020}, DRNet \cite{Schwab_2020}, and VCNet \cite{vcnet_2021}. We follow \cite{vcnet_2021} and apply target regularisation on DRNet and VCNet, and add a conditional density estimation head for DRNet. We also compare to statistical baselines such as GPS~\cite{hirano2004}, which is implemented using the \texttt{causal\_curve} package~\cite{kobrosly2020causal}, and EBCT~\cite{Tubbicke2022}, with VCNet employed as its inference network to estimate the ITRC. Finally, we test against causal forest double machine learning (DML)~\citep{chernozhukov2017doubledebiased} and non-parametric DML, which is available through the \texttt{EconML} package~\cite{econml}. 

We also benchmark CCPFN against top tabular foundation models (TabDPT~\cite{ma2025tabdpt}, TabPFN-2.6~\cite{grinsztajn2025tabpfn}, and TabICLv2~\cite{tabiclv2}) implemented as S-Learners~\cite{slearner}, which are not causally fine-tuned per dataset, but exhibit competitive performance (see Table \ref{tab:mise_results}). 

Crucially, practitioners lack access to ground-truth CEPOs during deployment, rendering baseline performance on benchmark datasets with respect to metrics such as MISE and DPE unsuitable for practical hyperparameter tuning. Therefore, during hyperparameter optimization, we select hyperparameters that minimize the training loss. The final reported performance is evaluated via 5-fold cross-validation on each test dataset. Further details on hyperparameter optimisation and our evaluation protocol are available in Appendix~\ref{sec:hpo} and \ref{sec:eval-protocol}.

\noindent
\textbf{Task 1: Estimation of treatment-response curves.} We first evaluate CCPFN and the baseline models on their ability to estimate a full individual treatment-response curve. The synthetic/semi-synthetic nature of our benchmarks gives us access to the full DGP that generates outcomes, thus allowing us to retrieve the ground-truth ITRC as well. We use the MISE metric \eqref{eq:mise_def} to evaluate performance on this task. 

\textbf{Task 2: Optimal policy prediction.} We also evaluate each model's ability to prescribe an optimal dosage/treatment level. Specifically, we want to test how well a model can predict the treatment $t^* = t^*(\textbf{x})$ that yields the optimal value of the treatment-response function $\mu_t(\textbf{x})$. We use the DPE metric \eqref{eq:dpe_def} to evaluate performance on this task. This metric is not applicable to all our benchmarks, specifically when the corresponding treatment-response functions are all monotonic, which occurs in the Debt dataset. Moreover, all tested models reported essentially zero DPE on MVICU. Thus we only report DPE for Warfarin, TCGA, News, and NewsHet. 

We perform ablations on the choice of prior and other design choices (such as the loss function, and whether or not to enforce positivity in the prior). We evaluate these in the context of both Task 1 and Task~2.

\vspace{-4pt}
\section{Results}
\label{sec:results}
\vspace{-4pt}

CCPFN demonstrates superior performance on the treatment-response curve estimation task across multiple benchmark datasets. When evaluated by average rank relative to mean integrated squared error (MISE), CCPFN consistently outperforms all baselines, achieving top results on the Debt and Warfarin datasets and coming in first place overall (Table \ref{tab:mise_results}).

\begin{table*}[ht]
\centering
    \setlength{\tabcolsep}{3pt}
\setlength{\extrarowheight}{-1pt}
\setlength{\aboverulesep}{0.5ex}
\setlength{\belowrulesep}{0.5ex}
\setlength{\cmidrulesep}{0.3ex}
\caption{Comparative evaluation of mean MISE across benchmark test datasets. Columns correspond to different benchmark datasets; values represent mean MISE $\pm$ standard deviation as computed with 5-fold cross-validation. First place is \textbf{bold}, second place is \underline{underlined}. Dashes (---) indicate no meaningful results were obtained. When evaluating TabPFN we apply PCA to reduce the dimension to 100, due to memory constraints and to match the dimensionality reduction used in CCPFN. DRNet, VCNet, and EBCT did not produce meaningful results for the MISE metric and hence are omitted.}
\small
\setlength{\tabcolsep}{6pt} 
\resizebox{\textwidth}{!}{%
\begin{tabular}{l | c | c c c c c c}
\toprule
\multirow{3}{*}{\textbf{Method}} & \multicolumn{7}{c}{\textbf{Mean MISE ($\downarrow$ better)}} \\
\cmidrule(lr){2-8}
& \makecell{MVICU \\ $(\times 10^3)$} & \makecell{Debt \\ $(\times 10^{-2})$} & Warfarin & TCGA & News & \makecell{NewsHet \\ $(\times 10^{-2})$} & Avg. Rank \\
\midrule
\textbf{CCPFN (Ours)} &\underline{ $1.45 \pm 2.4$} & $\mathbf{2.26 \pm .19}$ & $\mathbf{40.4 \pm 5.7}$ &  $8.63 \pm 1.8$ & $1.58 \pm .11$ & $5.76 \pm .39$ & \textbf{2.8} \\ 
\midrule
S-Learner (TabDPT)        & $1.53\pm2.5$ & \underline{$2.41\pm.22$} & \underline{$47.4\pm8.1$} & $8.47 \pm 1.3$ & \underline{$1.55 \pm .05$} & $6.28 \pm .16$ & \underline{3.3} \\
S-Learner (TabPFN)        & $1.59\pm 2.4$ & $5.80\pm .51$ & $159 \pm 37$ & $5.61 \pm .90$ & $\mathbf{1.49 \pm .06}$ & $6.33 \pm .11$ & 4.3 \\
S-Learner (TabICL)        & $1.49\pm2.5$ & $3.70 \pm .59$ & $54.8 \pm 11$ & \underline{$4.01 \pm .94$} & $1.68 \pm .05$ & $5.38 \pm .14$ &  3.5 \\
\midrule
ADMIT                     & $\mathbf{.250 \pm .06}$ & $2.64 \pm .33$ & $87.9 \pm 60$ & $6.21 \pm 1.7$  & $1.67 \pm .10$ & $8.12 \pm .32$ & 4.3 \\
SCIGAN                    & $2.61 \pm 2.2$ & $4.44 \pm .07$ & $711 \pm 44$ & $\mathbf{3.75 \pm .91}$ & $1.88 \pm .09$ & \underline{$4.72 \pm .41$} & 5.0 \\
GPS                       & $3.09 \pm 3.0$ & $16.8 \pm .32$ & $418 \pm 49$ & --- & $1.70 \pm .09$ & $6.36 \pm .13$ & 6.8 \\
CausalForest              & $70.7 \pm .00$ & $28.9 \pm 0.0$ & $894 \pm 24$ & $8.71 \pm 7.9$ & $1.78 \pm .57$ & $\mathbf{4.50 \pm .05}$ &  6.7  \\
NonparamDML               & $96.0 \pm .00$ & $39.4 \pm .57$ & $1230 \pm 49$ & $493 \pm 4.6$ & $1.63 \pm .39$ & $7.19 \pm .05$ & 7.8 \\
\bottomrule
\end{tabular}
}%
\label{tab:mise_results}
\end{table*}

Similarly, our model remains competitive on the DPE task (Table \ref{tab:dpe-results}). Notably, the performance of neural-network and statistical baselines varies significantly across datasets, suggesting a reliance on extensive hyperparameter tuning. On the other hand, CCPFN yields consistent results, despite the fact that the model \emph{does not see any evaluation data during pre-training}.

\begin{table*}[ht]
\centering
    \setlength{\tabcolsep}{3pt}
\setlength{\extrarowheight}{-1pt}
\setlength{\aboverulesep}{0.5ex}
\setlength{\belowrulesep}{0.5ex}
\setlength{\cmidrulesep}{0.3ex}
\caption{Comparative evaluation of mean DPE across benchmark test datasets. Columns correspond to different benchmark datasets; values represent mean DPE $\pm$ standard deviation as computed with 5-fold cross-validation. First place is \textbf{bold}, second place is \underline{underlined}. When evaluating TabPFN we apply PCA to reduce the dimension to 100, due to memory constraints and to match the dimensionality reduction used in CCPFN.}
\small
\setlength{\tabcolsep}{6pt} 
\resizebox{\textwidth}{!}{%
\begin{tabular}{l | c c c c c c}
\toprule
\multirow{2}{*}{\textbf{Method}} & \multicolumn{6}{c}{\textbf{Mean DPE ($\downarrow$ better)}} \\
\cmidrule(lr){2-7}
& Debt ($\times 10^{-3}$) & Warfarin & TCGA & News & NewsHet $(\times 10^{-3})$ & Avg. Rank \\
\midrule
\textbf{CCPFN (Ours)}     & $0.29 \pm .22$ & $2.91\pm1.7$ & $38.9 \pm 8.1$ & $3.71 \pm .60$ & $1.64 \pm .57$ & 6.0 \\ 
\hline
S-Learner (TabDPT)        & $8.24\pm3.2$ & $2.45 \pm 1.0$ & $35.5 \pm 8.2$ & $3.76 \pm .59$ & $0.74\pm .32$ & 5.0 \\
S-Learner (TabPFN)        & $72.5 \pm 13$ & $1.92\pm .82$ & $35.5 \pm 8.3$ & $\mathbf{2.69 \pm .30}$ & $1.14 \pm .48$ & 4.8 \\
S-Learner (TabICL)        & $14.4 \pm 6.2$ & \underline{$0.31 \pm .21$} & $35.8 \pm 6.7$ & $3.78 \pm .41$ & $\mathbf{.292 \pm .06}$ & \underline{4.6} \\
\hline
ADMIT                     & $0.26\pm .50$ & $\mathbf{0.10\pm .08}$ & \underline{$24.6 \pm 5.4$} & $3.65 \pm .67$ & \underline{$0.63\pm .04$} & \textbf{2.2} \\
SCIGAN                    & $\mathbf{0.00\pm .00}$ & $150\pm150$ & $38.6\pm 89$ & $3.88\pm .54$ & $6.91 \pm .57$ & 7.6 \\
GPS                      & $678\pm180$ & $80.9\pm2.6$ & --- & \underline{$3.46\pm .38$} & $1.04 \pm1.2$ & 6.3 \\
CausalForest              & $\mathbf{0.00 \pm .00}$ & $2010 \pm 24$ & $33.9 \pm 7.9$ & $4.37 \pm .57$ & $6.84 \pm .47$ & 7.2 \\
NonparamDML              & $28.2 \pm 5.7$ & $2000 \pm 49$ & $\mathbf{23.6 \pm 4.6}$ & $3.81 \pm .39$ & $6.84 \pm .47$ & 7.0 \\
DRNet                     & $64.7 \pm 70$ & $2.19\pm1.5$ & $29.5 \pm 6.6$ & $6.42\pm2.0$ & $5.35 \pm 4.6$ & 6.8 \\
VCNet                     & $207\pm140$ & $1800 \pm 1000$ & $31.6 \pm 12$ & $5.00\pm1.1$ & $6.84 \pm .47$ & 8.6 \\
EBCT                      & $327\pm240$ & $1400 \pm 1300$ & $45.0 \pm 12$ & $6.42\pm3.2$ & $1.91 \pm 2.8$ & 9.8 \\
\bottomrule
\end{tabular}
}
\label{tab:dpe-results}
\end{table*}

\begin{figure*}[ht!]
    \centering
    \begin{subfigure}[b]{0.48\textwidth}
        \centering
        \includegraphics[width=\textwidth]{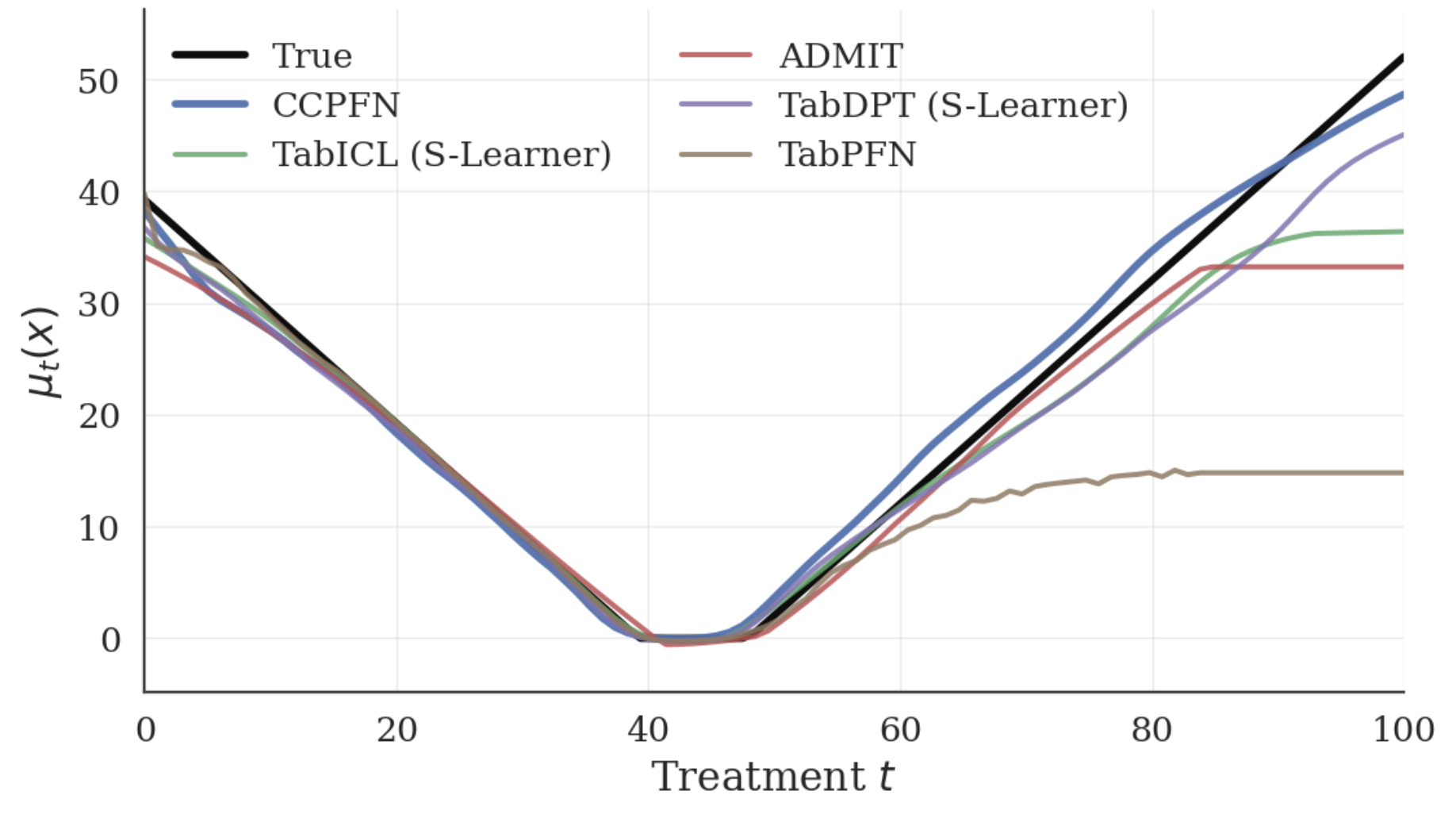}
        \caption{Individual A's ITRC }
        \label{fig:warfarin-itrc-1}
    \end{subfigure}
    \hfill
    \begin{subfigure}[b]{0.48\textwidth}
        \centering
        \includegraphics[width=\textwidth]{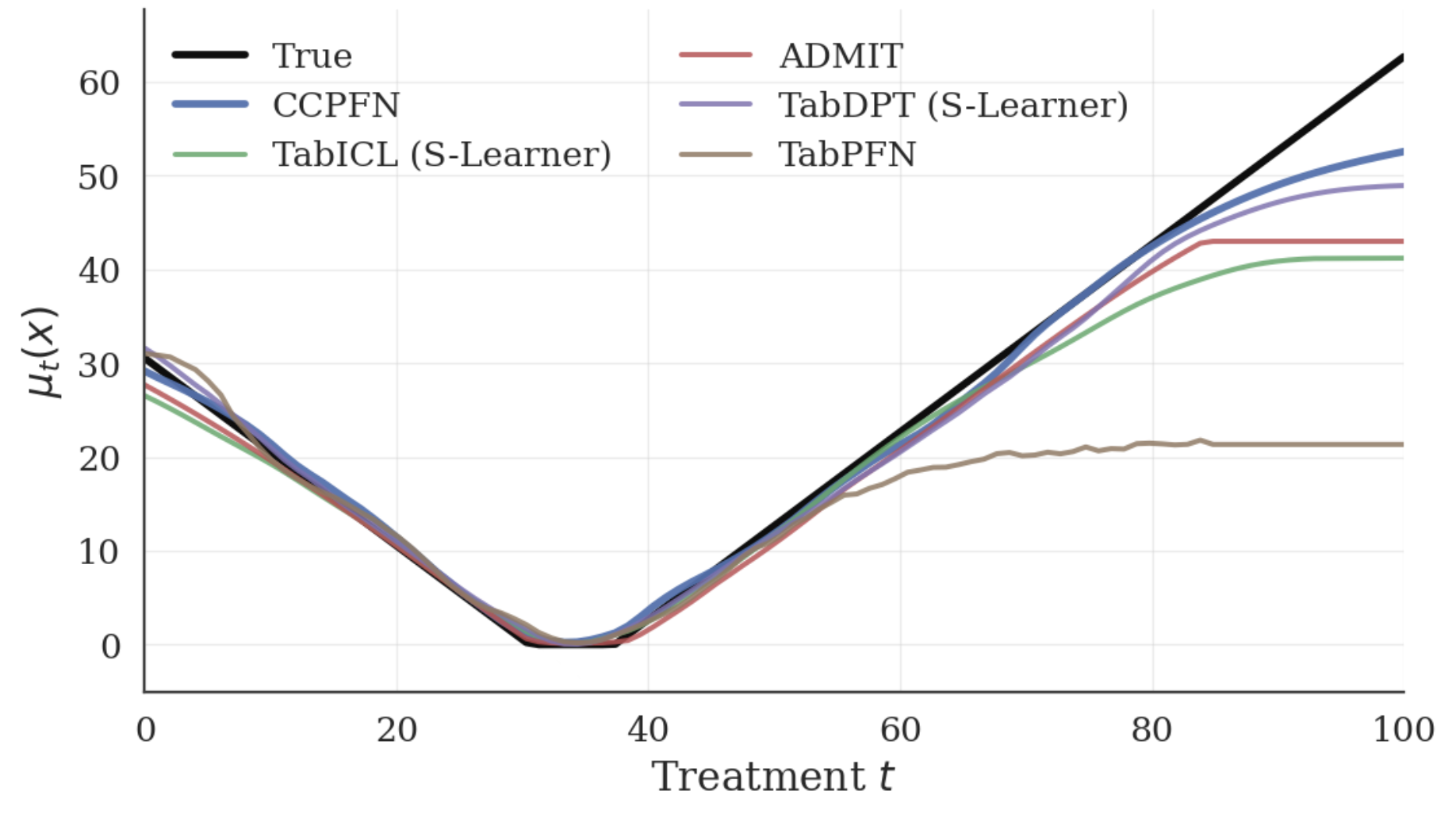}
        \caption{Individual B's ITRC}
        \label{fig:warfarin-itrc-2}
    \end{subfigure}
    \caption{Predicted individual treatment-response curves (ITRCs) and true ITRC for two randomly-selected individuals.}
    \label{fig:warfarin-itrc}
\end{figure*}

\textbf{ITRC Reconstruction} To qualitatively assess the model's ability to capture complex treatment-response dynamics, we visualize its predicted curves against the ground truth individual treatment-response curves. We focus on the Warfarin dataset, where the outcome represents the loss between a patient's administered dose and the optimal dose $t^*$, as determined by the IWPC pharmacogenetic dosing algorithm \cite{IWPC2009, kallus_zhou_2018}. The outcome is defined in terms of absolute distance from the optimal dosage $t^*$ for each individual, and a successful model must accurately recover its characteristic V-shape. Figure \ref{fig:warfarin-itrc} illustrates the predictions of the five top-performing methods. While all methods capture the general V-shape, CCPFN (shown in blue) tracks the ground truth more closely, particularly demonstrating superior accuracy near the endpoints of the dosage range.

\textbf{Ablations.} We perform ablations on important design choices related to our prior. We find a significant performance gain from including in-pass tabular ``corruption'' \cite{sui2024self, ma2025generalization} (e.g. binarizing, quantizing, or zero-inflating randomly-selected nodes) as opposed to solely post-hoc (Table \ref{tab:ablation-noinpass}). Next, we reconsider the necessity of imposing positivity in the prior for DGP generation. Positivity is one of the assumptions needed for identifiability of the treatment-response curve in theory. Indeed, we find that positivity is a necessary factor, and the prior without it generates data that is less effective for training CCPFN (\Cref{tab:ablation-nopos}). Further ablations can be found in Appendix \ref{sec:app-d}.

\begin{table*}[ht]
\centering
\caption{Ablation on applying tabular ``corruption'' during the data-generation process vs. only afterwards (post-hoc). Reported metrics are mean MISE and mean DPE (lower is better).}
\label{tab:ablation-noinpass}
\small

\setlength{\tabcolsep}{2.5pt}
\setlength{\extrarowheight}{-1pt}
\setlength{\aboverulesep}{0.5ex}
\setlength{\belowrulesep}{0.5ex}
\setlength{\cmidrulesep}{0.3ex}

\begin{tabular}{lcccccccccccc}
\toprule

& \multicolumn{2}{c}{\makecell{Debt \\ $(\times 10^{-2})$}}
& \multicolumn{2}{c}{\makecell{MVICU\\ $(\times 10^{-3})$}}
& \multicolumn{2}{c}{{Warfarin}}
& \multicolumn{2}{c}{{TCGA}}
& \multicolumn{2}{c}{{News}}
& \multicolumn{2}{c}{\makecell{NewsHet \\$(\times 10^{-2})$}} \\

\cmidrule(lr){2-3}
\cmidrule(lr){4-5}
\cmidrule(lr){6-7}
\cmidrule(lr){8-9}
\cmidrule(lr){10-11}
\cmidrule(lr){12-13}

\textbf{Method}
& \textbf{MISE}
& \textbf{DPE}
& \textbf{MISE}
& \textbf{DPE}
& \textbf{MISE}
& \textbf{DPE}
& \textbf{MISE}
& \textbf{DPE}
& \textbf{MISE}
& \textbf{DPE}
& \textbf{MISE}
& \textbf{DPE}
\\

\midrule

\textbf{Ours}
& \textbf{2.22}
& \textbf{0.011}
& \textbf{1.45}
& --
& \textbf{36.6}
& \textbf{2.62}
& \textbf{8.63}
& 38.9
& 1.57
& 3.80
& \textbf{5.58}
& \textbf{0.152}
\\

Post-Hoc Corruption
& 3.75
& 0.311
& 1.53
& --
& 54
& 2.66
& 9.11
& \textbf{36.5}
& \textbf{1.53}
& \textbf{3.22}
& 6.24
& 0.252
\\
\bottomrule
\end{tabular}
\end{table*}

\begin{table*}[ht]
\centering
\caption{Ablation on enforcing positivity in the prior on DGPs. No positivity enforced entails having $\mlpt$ output the observed $T$ directly, with no noise node $\eta_T$ added. Reported metrics are mean MISE and mean DPE (lower is better).}
\label{tab:ablation-nopos}
\small

\setlength{\tabcolsep}{2.5pt}
\setlength{\extrarowheight}{-1pt}
\setlength{\aboverulesep}{0.5ex}
\setlength{\belowrulesep}{0.5ex}
\setlength{\cmidrulesep}{0.3ex}

\begin{tabular}{lcccccccccccc}
\toprule

& \multicolumn{2}{c}{\makecell{Debt \\ $(\times 10^{-2})$}}
& \multicolumn{2}{c}{\makecell{MVICU \\ $(\times 10^{-3})$}}
& \multicolumn{2}{c}{{Warfarin}}
& \multicolumn{2}{c}{{TCGA}}
& \multicolumn{2}{c}{{News}}
& \multicolumn{2}{c}{\makecell{NewsHet \\$(\times 10^{-2})$}} \\

\cmidrule(lr){2-3}
\cmidrule(lr){4-5}
\cmidrule(lr){6-7}
\cmidrule(lr){8-9}
\cmidrule(lr){10-11}
\cmidrule(lr){12-13}

\textbf{Method}
& \textbf{MISE}
& \textbf{DPE}
& \textbf{MISE}
& \textbf{DPE}
& \textbf{MISE}
& \textbf{DPE}
& \textbf{MISE}
& \textbf{DPE}
& \textbf{MISE}
& \textbf{DPE}
& \textbf{MISE}
& \textbf{DPE}
\\

\midrule

\textbf{Ours}
& \textbf{2.22}
& \textbf{0.011}
& \textbf{1.45}
& --
& \textbf{36.6}
& \textbf{2.62}
& 8.63
& 38.9
& \textbf{1.57}
& \textbf{3.80}
& \textbf{5.58}
& \textbf{0.152}
\\

No Positivity Enforced
& 9.98
& 88.6
& 1.48
& --
& 39.7
& 3.34
& \textbf{7.11}
& \textbf{36.0}
& 1.68
& 4.07
& 6.97
& 0.316
\\

\bottomrule
\end{tabular}
\end{table*}

\vspace{-4pt}
\section{Conclusion, Limitations, and Future Work}
\label{sec:conclusion}
\vspace{-4pt}

We introduce CCPFN, a causal foundation model for the continuous treatment setting. It demonstrates superior ability to reconstruct continuous individual treatment-response curves without any further fine-tuning on unseen datasets. A central contribution is our novel 3-MLP prior which naturally generates causal DGPs satisfying unconfoundedness and positivity. 

We are limited by the lack of broad, large-scale causal evaluation datasets for the continuous treatment setting. Moreover, CCPFN relies on the unconfoundedness and positivity assumptions. Unconfoundedness is impossible to verify in practice, and domain experts must be consulted to assess whether or not it is valid. Meanwhile, in the continuous treatment setting, it is impossible for positivity to hold exactly, as the possible treatment range is uncountably infinite. We are also limited by our embedding dimension, as datasets with large features are compressed into lower dimensional representations.

There are many possible directions of future research. In regions of the treatment range $\mathcal{T}$ where observational data is scarce, we expect the model to have larger epistemic uncertainty regarding the ITRC values. As CCPFN already predicts the full CEPO-PPD, this is a natural next step. It would also be of interest to extend to the case of multi-arm treatments as in \cite{Schwab_2020}.

\newpage
\bibliography{bib}


\appendix
\newpage
\section{Further Details About Benchmarks and Baselines}
\label{sec:app-a}

\subsection{Synthetic and Semi-Synthetic Data Scenarios}\label{sec:scenarios}

Each subclass of \texttt{Scenario} at minimum implements the following three methods:
\begin{itemize}[noitemsep,topsep=0pt,leftmargin=*]
    \item \texttt{load\_covariates}: Generates the base covariates $\mathcal{D}_{\text{cov}} = \{\mathbf{x}_n\}_{n = 1}^N$. For semi-synthetic data, these are loaded from real data; for fully synthetic data, these are generated synthetically.
    \item \texttt{treatment}: Generates the treatment administered to each individual.
    \item \texttt{dose\_response}: The noise-free treatment-response function $(\mathbf{x}, t) \mapsto \mu_t(\mathbf{x})$ for this \texttt{Scenario}.
\end{itemize}
Example plots of treatment-response curves generated by each scenario can be found in Appendix \ref{sec:dataset-details} (test datasets) and Appendix \ref{sec:validation-data-construction} (validation datasets).

\subsection{Benchmark Details}\label{sec:dataset-details}

\begin{figure}[htbp]
    \centering
    
    \begin{subfigure}[b]{0.45\textwidth}
        \centering
        \includegraphics[width=\linewidth]{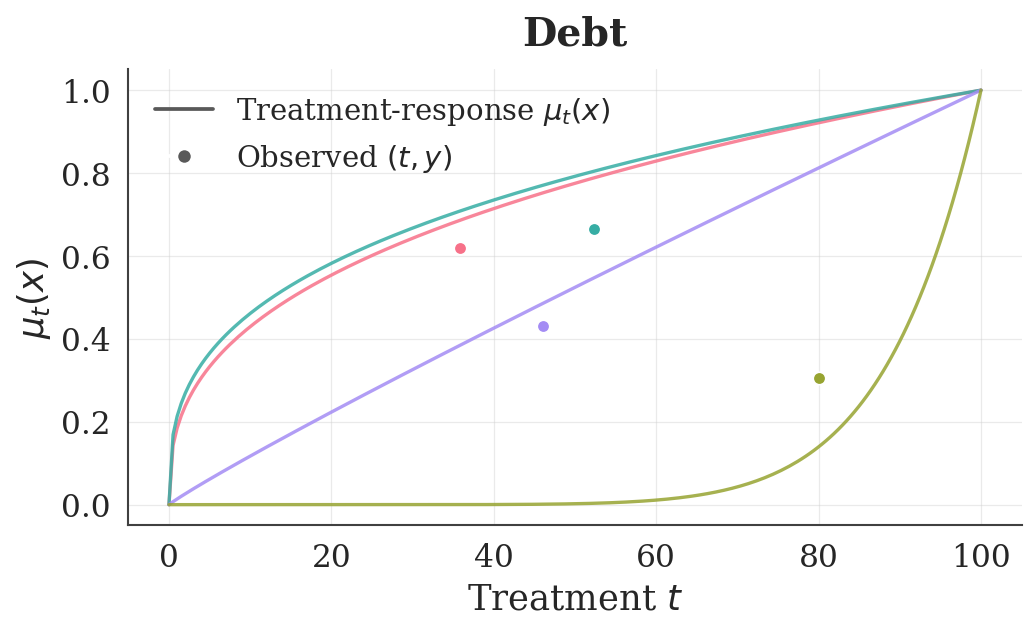}
    \end{subfigure}
    \hfill
    \begin{subfigure}[b]{0.45\textwidth}
        \centering
        \includegraphics[width=\linewidth]{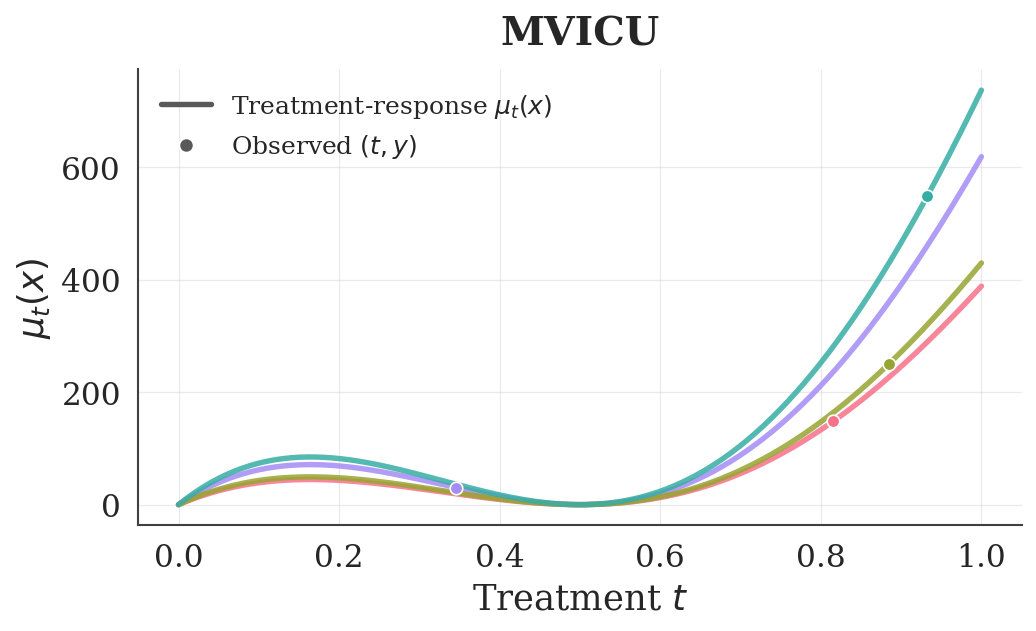}
    \end{subfigure}
    
    \vspace{0.3cm} 

    \begin{subfigure}[b]{0.45\textwidth}
        \centering
        \includegraphics[width=\linewidth]{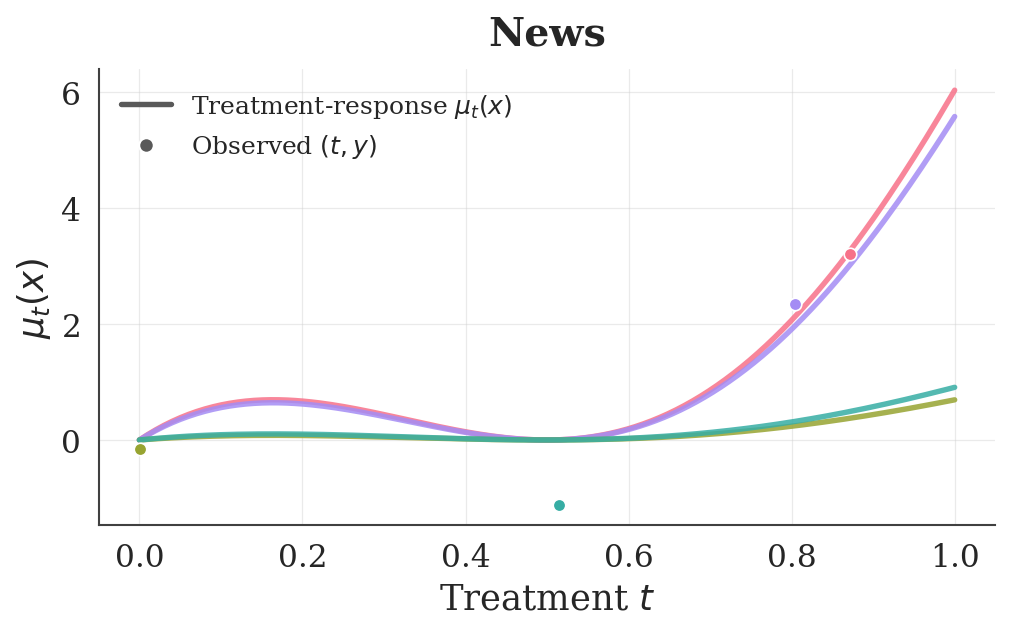}
    \end{subfigure}
    \hfill
    \begin{subfigure}[b]{0.45\textwidth}
        \centering
        \includegraphics[width=\linewidth]{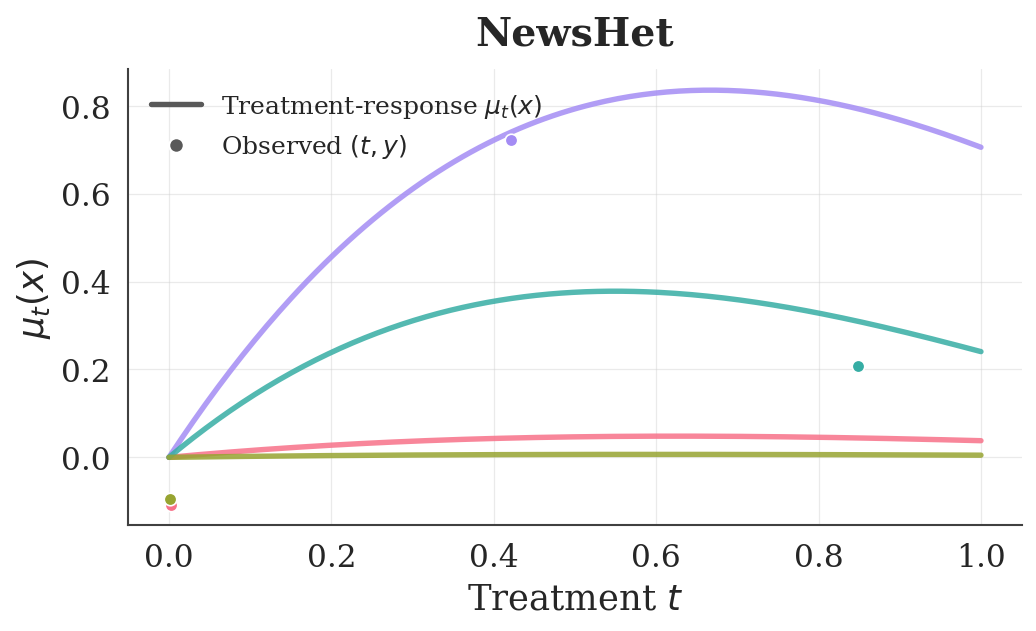}
    \end{subfigure}
    
    \vspace{0.3cm}

    \begin{subfigure}[b]{0.45\textwidth}
        \centering
        \includegraphics[width=\linewidth]{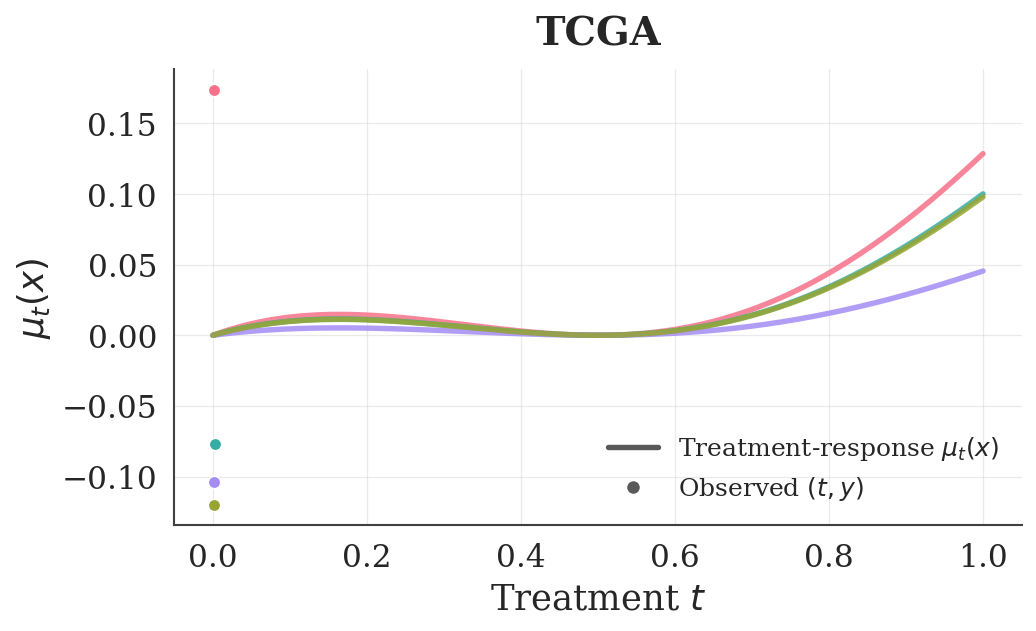}
    \end{subfigure}
    \hfill
    \begin{subfigure}[b]{0.45\textwidth}
        \centering
        \includegraphics[width=\linewidth]{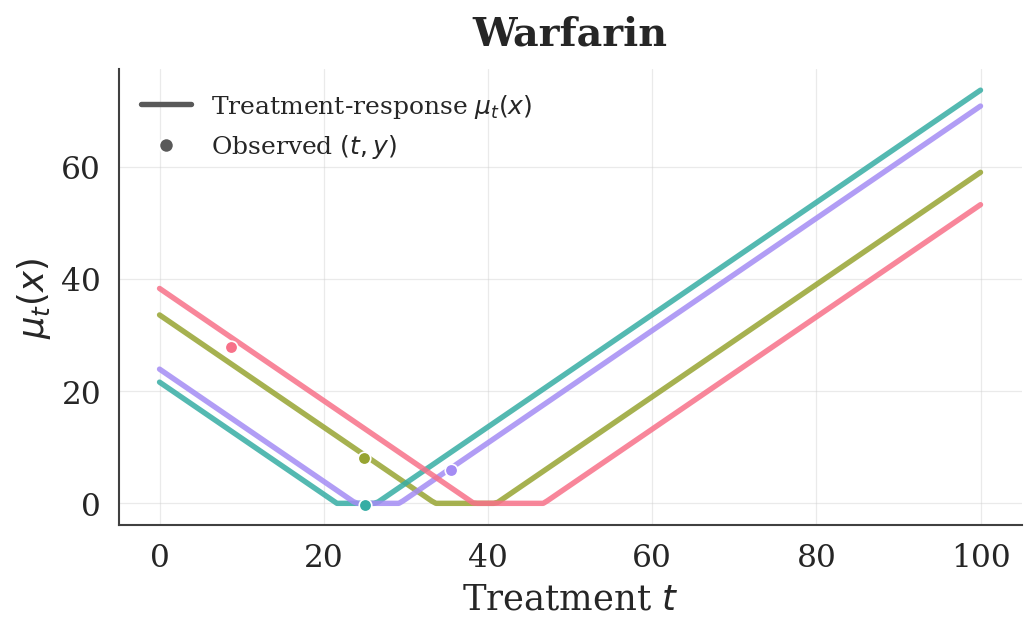}
    \end{subfigure}
    
    \caption{Example individual treatment-response curves (ITRCs) for all six test scenarios. Solid curves are ITRCs for randomly-selected individuals; circles are the corresponding observed $(T, Y)$. Note that the observations do not lie precisely on the ITRCs due to the presence of exogenous noise; the curves represent expectations (CEPOs), not exact counterfactual outcomes.}
    \label{fig:combined_plots_all_test}

\end{figure}

We used 6 different held-out datasets for testing CCPFN against other causal inference methods. For each set of covariates, a dataset provides a factual treatment and outcome, as well as counterfactual treatments along with their corresponding CEPO.
\begin{itemize}[noitemsep,topsep=0pt,leftmargin=*]
    \item \textbf{MVICU.} A semi-synthetic medical scenario adapted from \cite{Schwab_2020} and post-processed. It models the effect of different configurations of mechanical ventilation in the intensive care unit on patients. For treatment and outcome functions, we used the same ones as in News \cite{admit_2022}. It contains 13 covariates and 4,963 rows. (PhysioNet Credentialed Health Data License 1.5.0)
    \item \textbf{Debt.} A fully synthetic financial scenario concerning the impact of different levels of debt write-downs on total debt repaid, adapted from \cite{causal-pipeline-2025}. The generated dataset has 10 covariates and 10,000 rows. (MIT license)
    \item \textbf{News.} A semi-synthetic scenario representing reader interaction with news articles from \cite{Schwab_2020, admit_2022}. It has 2,870 covariates. Using the original dataset downloaded from \cite{Schwab_2020}, we use the process outlined in \cite{admit_2022} to process the data. Specifically, we randomly subsampled 10,000 rows. From this, we further subsampled 7,881 by dropping rows whose optimal CEPO (as determined by the oracle) was greater than 10. (MIT license)
    \item  \textbf{NewsHet.} A semi-synthetic scenario adapted from News, with treatment-response functions modified to introduce heterogeneity of optimal dosage levels. As implemented in \cite{admit_2022}, News suffers from the fact that all optima occur at the endpoints 0 or 1, making the optimal dosage policy prediction a degenerate task We introduce heteroscedastic horizontal shifts to rectify this. (MIT license)
    \item \textbf{TCGA.} A semi-synthetic scenario representing the effect of medication dosage and risk of cancer recurrence from \cite{Schwab_2020, admit_2022}, where we use the DGP outlined by \cite{admit_2022} to generate ground-truth causal effects. We additionally removed rows whose optimal CEPO was greater than 50, leaving 4,428 rows. (MIT license)
    \item \textbf{Warfarin.} A semi-synthetic medical scenario concerning warfarin dosing, adapted from \cite{IWPC2009, warfarin, kallus_zhou_2018}. The dataset specifically calculates the loss between the actual dosage received by an individual and the optimal warfarin dosage calculated by the IWPC pharmacogenetic algorithm \cite{IWPC2009} and uses this as the outcome. The generated dataset has 19 covariates and 4,490 rows. (CC BY-SA 4.0)
\end{itemize}
Example individual treatment-response curves are shown for each benchmark dataset in Figure \ref{fig:combined_plots_all_test}.

\subsection{Hyperparameter Optimization}\label{sec:hpo}

\textbf{EconML Baselines}  We tune results from the EconML Baselines (CausalForestDML and NonParamDML) using the FLAML (AutoML) library. Hyperparameter tuning is performed on the treatment model and the outcome model. We use $k$-fold cross-validation with $k = 5$, early stopping, and a time budget of 900 seconds. The following base estimators are used: \texttt{"lgbm", "xgboost", "xgb\_limitdepth", "rf", "kneighbour", "extra\_tree"}. 

\textbf{Neural Network Methods} For the neural network baselines, we conduct a grid search over learning rates $\in \{0.005, 0.001, 0.0005, 0.0003, 0.0001\}$ and batch sizes $\in \{128, 256, 512, 1024\}$. For each dataset, we sample 80\% of the data and perform 5-fold cross-validation on this sample to identify the hyperparameter configuration yielding the lowest average training loss for each baseline. We use the training loss because in practice, one \emph{never} has access to the synthetic validation metrics used. Once the optimal hyperparameters are identified, we perform 5-fold cross-validation on the entire training dataset to elicit final performance. 

\subsection{Evaluation Protocol}\label{sec:eval-protocol}
The fully optimized versions of all considered models (CCPFN, TFMs, neural networks, EconML, and statistical baselines) are evaluated on the benchmark datasets using $5$-fold cross-validation. For the neural network, statistical, and EconML baselines, the training splits are used to explicitly fit the models to each benchmark scenario. Conversely, for CCPFN and TFMs, these training splits are provided solely as in-context examples, leaving the models' underlying weights frozen. The final reported MISE and DPE metrics for each method are the mean error across the five validation folds as well as the standard deviation across these five folds.

\section{Further Detail on Synthetic Priors}
\label{sec:app-b}
\subsection{Our 3-MLP Prior}

During training, the number of samples $N$ is fixed to be 2048. The maximum number of covariates is 98. The number of layers $L_{\mathbf{X}}, L_T, L_Y$ are all sampled from independent truncated normal distributions with mean and variance $\alpha > 0$, where $\alpha \sim \text{LogUniform}(A, B)$. In our experiments, we set $A = 1, B = 10$ for $\mathbf{X}, T$, and $Y$. The number of hidden units $H_{\mathbf{X}}, H_T, H_Y$ are also sampled from the same class of distributions, with $A = 10, B = 100$ for $\mathbf{X}, T$, and $Y$. The noise scale is also sampled from this class of distributions, with $A = 1 \times 10^{-4}$ and $B = 0.5$. The densities $d_{\mathbf{X}}, d_T, d_Y$ were sampled independently from $\mathcal{U}(0.1, 1)$. To ensure that the signal from $T$ is not lost when dropping edges, we protect all outgoing edges from $T$ in the input layer of $\mlpy$.

In the truncated normal distributions, we truncate to $[3, \infty)$ for layer distributions and $[4, \infty)$ for hidden size distributions. Integer values are then obtained by rounding.

\begin{figure}[htbp]
    \centering
    \includegraphics[width=1\linewidth]{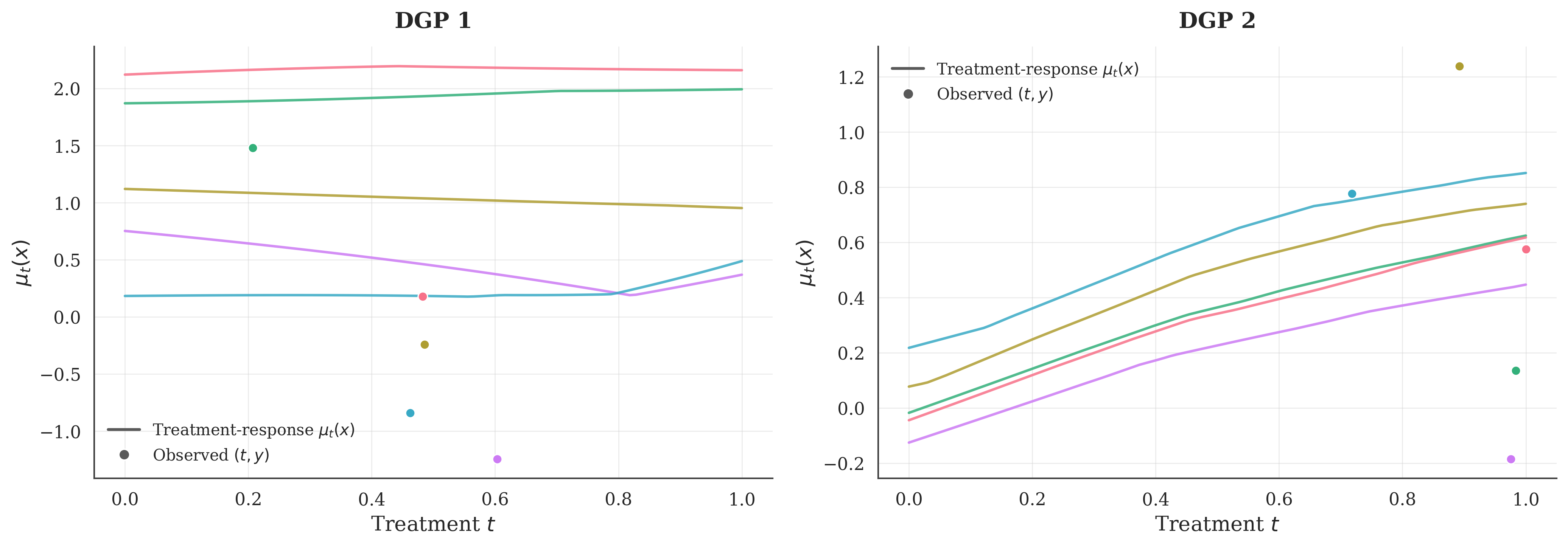}
    \caption{Example individual treatment-response curves (ITRCs) from different DGPs produced by our 3-MLP prior. Solid curves are ITRCs for randomly-selected individuals; circles are the corresponding observed $(T, Y)$. Note that the observations do not lie precisely on the ITRCs due to the presence of exogenous noise; the curves represent expectations (CEPOs), not exact counterfactual outcomes.}
    \label{fig:placeholder}
\end{figure}

\subsection{Alternative Priors}\label{app:alt-prior}

We compare two alternative prior distributions over data-generating processes.

\paragraph{Bernstein polynomial prior.}
We sample a table of $N$ rows and partition its columns into covariates $\mathbf{X}$ and a noise column $\boldsymbol{\epsilon}$. Covariates are standardised, and treatments are drawn from a sigmoid-normal distribution,
\begin{equation}
    t \mid \mathbf{x} \sim \sigma\!\left(\mathcal{N}(\mu_{t \mid \mathbf{x}},\, \sigma^2_{t \mid \mathbf{x}})\right),
\end{equation}
where $\mu_{t \mid \mathbf{x}}$ and $\sigma_{t \mid \mathbf{x}}$ are the outputs of a conditional MLP applied row-wise to $\mathbf{x}$. Overlap is controlled by scaling $\sigma_{t \mid \mathbf{x}}$ by an overlap parameter $\alpha \in (0,1]$. The conditional expected potential outcome (CEPO) $\mu_t(\mathbf{x})$ is defined via a $K$-degree Bernstein polynomial, whose coefficients are formed as a convex combination of individual-specific coefficients $\mathbf{c}(\mathbf{x})$ (produced by a second conditional MLP) and a shared global coefficient vector $\mathbf{c}_0 \sim \mathcal{N}(\mathbf{0}, \mathbf{I})$:
\begin{equation}
    \mathbf{c} = \lambda\, \mathbf{c}(\mathbf{x}) + (1 - \lambda)\, \mathbf{c}_0,
\end{equation}
where $\lambda \in [0,1]$ governs the degree of treatment effect heterogeneity. The observed outcome is then $y = \mu_t(\mathbf{x}) + \epsilon$, where $\epsilon$ is a scaled, centred noise term derived from the reserved noise column.

\paragraph{Value-based prior.}
We again sample a table of $N$ rows and select a subset of columns as covariates $\mathbf{X}$. Rather than parameterising the treatment-response curve analytically, we directly read off potential outcomes at $n$ randomly sampled (and sorted) treatment support points $\{t_1, \ldots, t_n\} \subset [0,1]$. For each support point $t_k$, we reserve one table column as the CEPO $\mu_{t_k}(\mathbf{x})$ and a second column as an individual-level noise term $\eta_k(\mathbf{x})$; noise columns are scaled so that their contribution is a fixed fraction of the corresponding signal variance. Observed outcomes at any treatment value $t$ are obtained by linearly interpolating both $\mu_{t_k}(\mathbf{x})$ and $\eta_k(\mathbf{x})$ across the two nearest support points and summing the results. Treatments are sampled using the same sigmoid-normal mechanism as in the Bernstein prior.

\section{Model Training Details}
\label{sec:app-c}

\subsection{Model Architecture and Hyperparameters}\label{sec:arch-hypers}

We use a modified version of the TabDPT architecture \cite{ma2025tabdpt} with a nonlinear $T$-encoder (see Figure \ref{fig:model-arch}). We initialize our weights to TabDPT's trained weights on layers which support them.
When the number of covariates $K$ exceeds 100, we apply truncated SVD to the covariates before passing them to the model. During training, we used a batch size of 32, 8 gradient accumulation steps, and 128 model updates per epoch. Our final model was fine-tuned for 15 epochs; this number was selected by early stopping based on the validation dataset results. The $\sigma$ used in the histogram loss \eqref{eq:histogram-loss} is $0.01$. We use 20 transformer layers with 6 heads per layer, an embedding dimension of 384, and a feed-forward hidden dimension of 768.

The final model architecture has 19,140,352 parameters. Of these, 297,600 are the nonlinear $T$-encoder parameters, compared to 39,168 for the linear $\mathbf{X}$-and-$T$-encoder and 768 for the $Y$-encoder.

All training runs were performed on single NVIDIA A6000 GPUs.

\subsection{Validation Dataset Construction}\label{sec:validation-data-construction}

The datasets we used for model validation and hyperparameter selection were a mix of synthetic and semi-synthetic. We used the fully synthetic dataset constructed in \cite{admit_2022}, as well as a simple linear dataset with a linear treatment-response function. We chose the best version of our model based on averaged IQR-normalised MISE performance across all validation datasets. 

\begin{figure}[htbp]
    \centering
    \begin{subfigure}[b]{0.45\textwidth}
        \centering
        \includegraphics[width=\textwidth]{content//images/acic2016-plot.png}
        \label{fig:acic2016-plot.png}
    \end{subfigure}
    \hfill
    \begin{subfigure}[b]{0.45\textwidth}
        \centering
        \includegraphics[width=\textwidth]{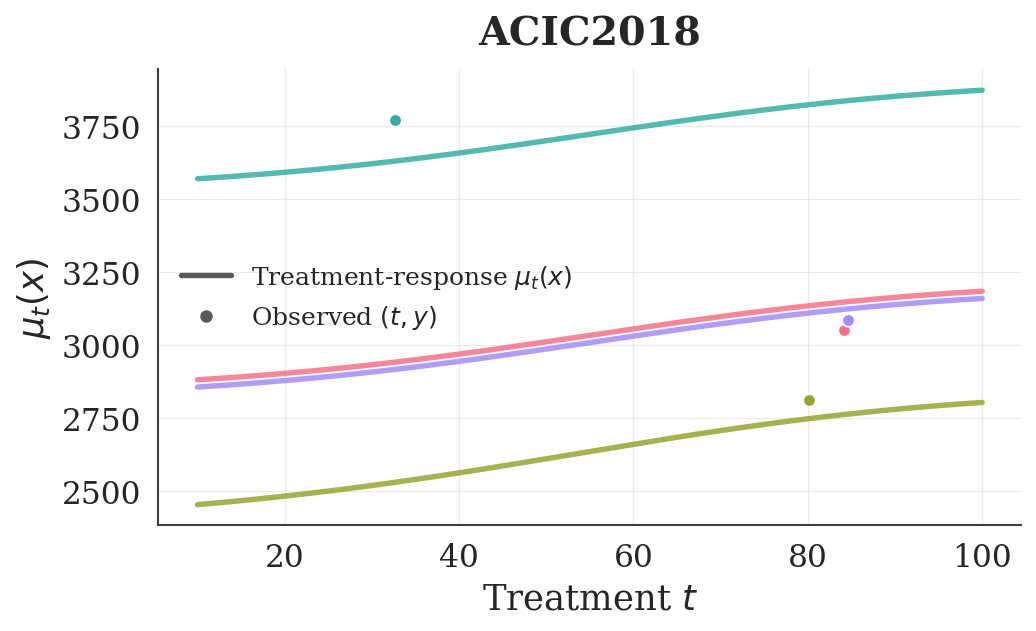}
        \label{fig:acic2018-plot.png}
    \end{subfigure}
    \vskip\baselineskip
    
    \begin{subfigure}[b]{0.45\textwidth}
        \centering
        \includegraphics[width=\textwidth]{content//images/criteo-plot.png}
        \label{fig:criteo-plot.png}
    \end{subfigure}
    \hfill
    \begin{subfigure}[b]{0.45\textwidth}
        \centering
        \includegraphics[width=\textwidth]{content//images/hillstrom-plot.png}
        \label{fig:hillstrom-plot.png}
    \end{subfigure}
    \vskip\baselineskip
    
    \begin{subfigure}[b]{0.45\textwidth}
        \centering
        \includegraphics[width=\textwidth]{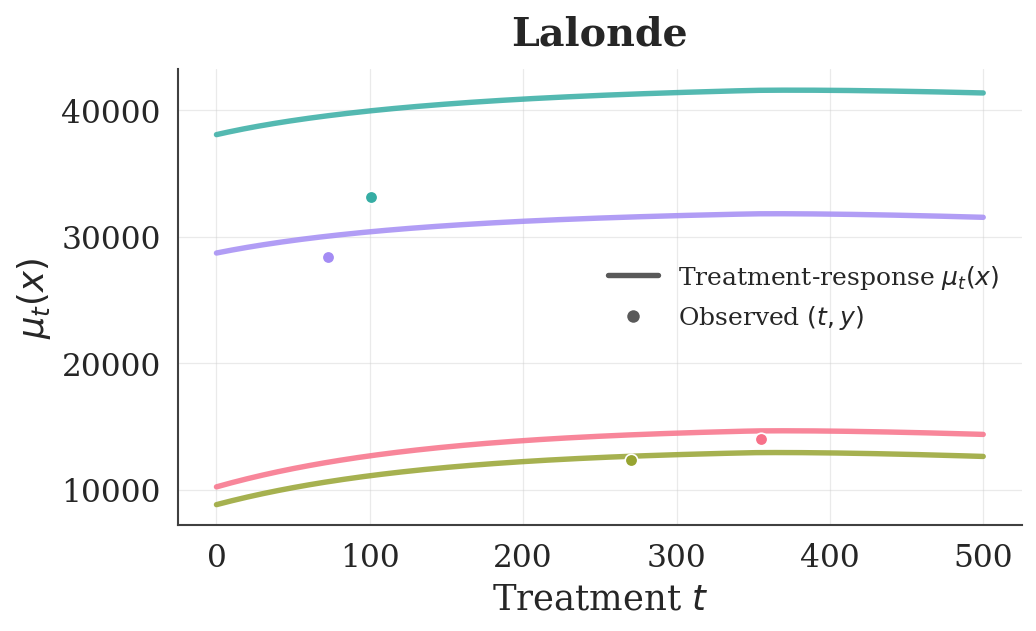}
        \label{fig:lalonde-plot.png}
    \end{subfigure}
    \hfill
    \begin{subfigure}[b]{0.45\textwidth}
        \centering
        \includegraphics[width=\textwidth]{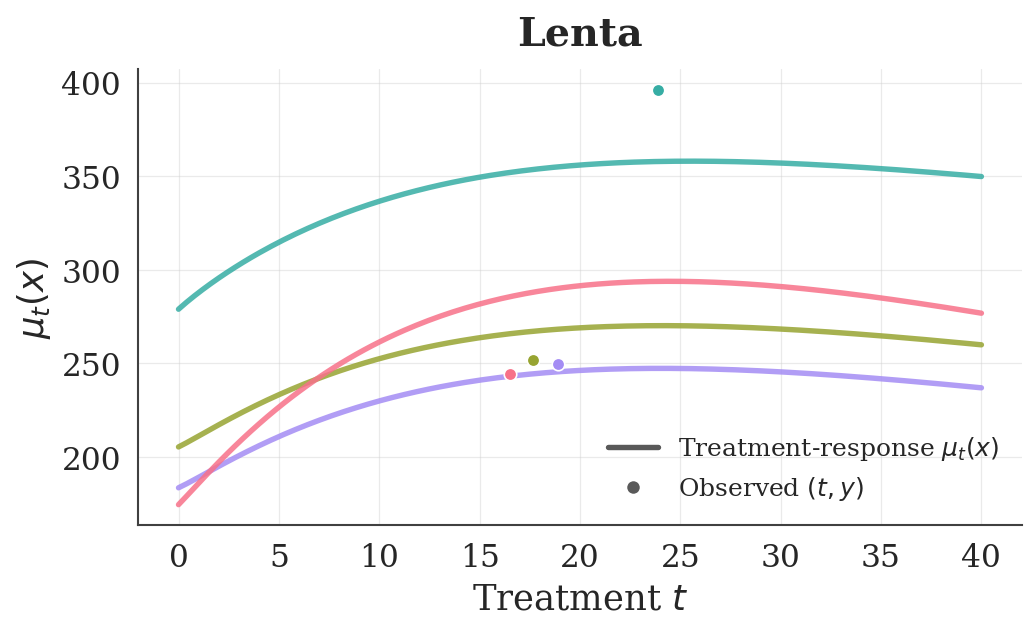}
        \label{fig:lenta-plot.png}
    \end{subfigure}
    \vskip\baselineskip
    
    \begin{subfigure}[b]{0.45\textwidth}
        \centering
        \includegraphics[width=\textwidth]{content//images/twins-plot.png}
        \label{fig:twins-plot.png}
    \end{subfigure}
    \hfill
    \begin{subfigure}[b]{0.45\textwidth}
        \centering
        \includegraphics[width=\textwidth]{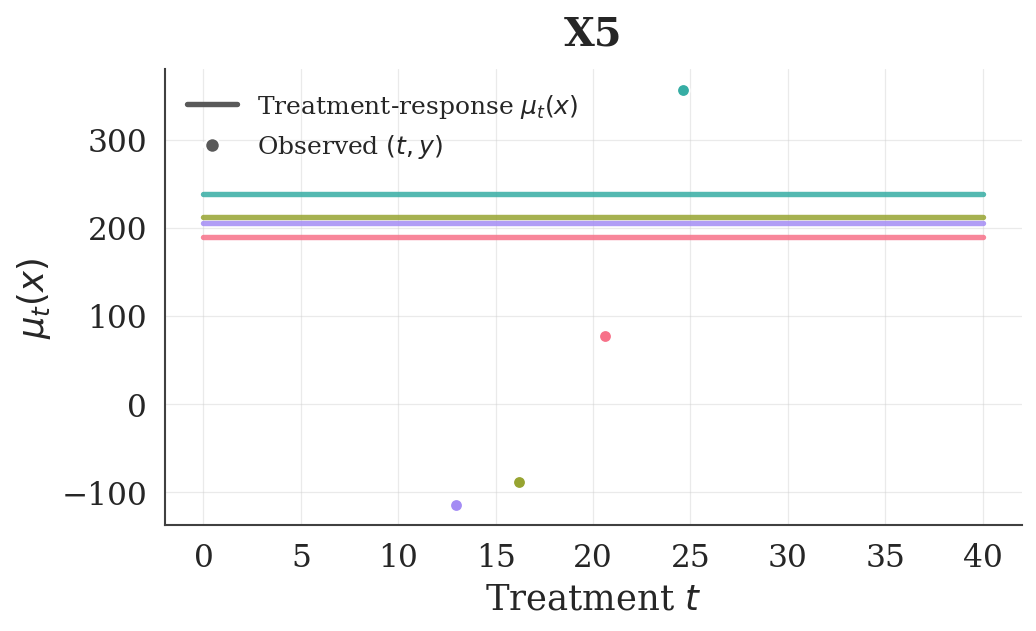}
        \label{fig:x5-plot.png}
    \end{subfigure}
    
    \caption{Example individual treatment-response curves (ITRCs) for all eight validation scenarios. Solid curves are ITRCs for randomly-selected individuals; circles are the corresponding observed $(T, Y)$. Note that the observations do not lie precisely on the ITRCs due to the presence of exogenous noise; the curves represent expectations (CEPOs), not exact counterfactual outcomes.}
    \label{fig:combined-plots-validation-all}
\end{figure}

For semi-synthetic data, we used covariates from several well-known causal inference datasets: ACIC2016 (Apache license) \cite{acic2016}, ACIC2018 (Apache license) \cite{acic2018}, Criteo (MIT license) \cite{criteo}, Hillstrom (MIT license) \cite{hillstrom2008}, Lalonde (MIT License) \cite{dehejia1999causal, causaldata_py}, Lenta (MIT license) \cite{lenta2020bigtarget, shevchenko2020scikituplift}, Twins (MIT license) \cite{realcause}, and X5 (MIT license) \cite{x5retailhero2019, shevchenko2020scikituplift}. We tasked an LLM agent with generating a plausible scenario for each set of covariates using the prompts included below. Example individual treatment-response curves for each of the validation datasets are shown in Figure \ref{fig:combined-plots-validation-all}.

\begin{tcolorbox}[
    title=System Prompt for Generating Synthetic Validation Data \#1,
    colback=gray!10,
    colframe=gray!70,
    coltitle=black,
    fonttitle=\bfseries,
    breakable,
    enhanced,
    bottomrule at break=0pt,
    toprule at break=0pt
]

\begin{lstlisting}[breaklines=true, basicstyle=\ttfamily\footnotesize, columns=fullflexible]
# Semi-synthetic data generation instructions

(Note: run python scripts in the `conda tracee` environment.)

## Background
You are working on a project in causal inference. The goal is to train a model to perform causal inference in the *continuous treatment* setting.

## Your Task
Your task is to create semi-synthetic data consisting of real-world covariates `X` (real data) and synthetic treatment and outcome variables by creating synthetic data-generating processes (DGPs). Adhere to the following instructions:

1. Ask the user which `csv` file to use as the base covariates `X`. It is possible that there is no local csv, and the covariates will have to be downloaded in the script itself (e.g. using sklearn.datasets). You can download and view the covariates now, so that you have intuition for the context.
2. Ask the user for covariate context, i.e. what do the base covariates `X` represent in this dataset?
3. Ask the user for treatment and outcome context, i.e. what scenario the user has in mind for the treatment and outcomes. 
4. Based on the information provided in steps 1 - 3, devise a *realistic* DGP to simulate treatment assignment and outcomes. Remember, the treatment variable should be continuous, *not* binary. This DGP should satisfy the following requirements:
  1. There should be a high degree of confounding: at least 50% of the covariates should be causes of both the treatment and the outcome. 
  2. You should generate a *dose-response function* f(X, t) that maps an individual with covariates X and hypothetical treatment t to the *conditional expected potential outcome (CEPO)*. This should be a suitably complex and realistic function which can be implemented in simple Python code. 
  3. You should generate a *treatment assignment function* T(X) that maps an individual with covariates X to the *observed* treatment T(X). This should be a suitably complex and realistic function which can be implemented in simple Python code. 
  4. In order to ensure that there is a high degree of confounding, the functions f and T should both depend on some subset of covariates comprising at least half of the total number of covariate features. 
5. Once you have constructed this DGP, generate a *Python script* that outputs a csv file as follows:
  1. Ask the user for the desired name of the Python script.
  2. The Python script should include code for the dose-response function f(X, t) and the treatment assignment function T(X). 
  3. The Python script should output a single csv file with columns named x_0 through x_n (where n is the number of covariate features), t, y, t_test, cepo_test. The data should be filled as follows: 
    1. The values of columns x_0 through x_n should be the values of the original base covariates csv. 
    2. The value of t should be the value of T(X) for X the corresponding covariate value.
    3. The value of y should be f(X, t) for X the corresponding covariate value and t = T(X), *plus Gaussian noise* which is iid for each row. 
    4. The value of t_test should be randomly sampled from [t_min, t_max]. 
    5. The value of cepo_test should be f(X, t_test).
    6. All data should be numerical (e.g. string-based categorical variables should be encoded as integers).
  4. Save the python script in tracee/inference/benchmarks/data_generation_scripts. 

When you are ready to proceed with this task, begin at step 1 above.
    
\end{lstlisting}

\end{tcolorbox}

\begin{tcolorbox}[
    title=System Prompt for Generating Synthetic Validation Data \#2,
    colback=gray!10,
    colframe=gray!70,
    coltitle=black,
    fonttitle=\bfseries,
    breakable,
    enhanced,
    bottomrule at break=0pt,
    toprule at break=0pt
]

\begin{lstlisting}[breaklines=true, basicstyle=\ttfamily\footnotesize, columns=fullflexible]

#Semi-synthetic data generation instructions 

##Background 

You are working on a project in causal inference. The goal is to train a model to perform causal inference in the continuous treatment setting. ##Your Task Your task is to create semi-synthetic data consisting of real-world covariates X (real data) and synthetic treatment and outcome variables by creating synthetic data-generating processes (DGPs). The dataset we are working with is the Lalonde dataset. You can decide the best way to access this dataset (realcause, jobs etc.) Only use the base covariates. Propose a potential CONTINUOUS-VALUED treatment (T) and CONTINUOUS-VALUED outcome (Y). You should then realise a DGP to generate this treatment and outcome. Some things to keep in mind are:

We are working in the backdoor causal graph scenario, where X -> T, T -> Y, and X-> Y

There should be a high degree of confounding - at least 50% of the covariates should affect outcome Y You should generate a dose-response function f(X, t) that maps an individual with covariates X and hypothetical treatment t to the conditional expected potential outcome (CEPO). This should be a suitably complex and realistic function which can be implemented in simple Python code. You should generate a treatment assignment function T(X) that maps an individual with covariates X to the observed treatment T(X). This should be a suitably complex and realistic function which can be implemented in simple Python code. This DGP should be generated in Python code. You should then save the data into a csv file as follows, where columns are X_0, ..., X_n, t, y, t_test, cepo_test. n refers to the number of covariates we originally had. Each row represents one individual in the original lenta dataset

Rows X_0, X_n should retain the base covariate values from the original dataset

t = T(X) for corresponding X value (where X represents the set of covariates)

y = f(X, t) for X the corresponding covariate value and t + Gaussian noise which is iid for each row.

t_test should be randomly sampled from a suitable range based on the nature of the continuous treatment you propose

cepo_test = f(X, t_test) for the corresponding t_test This should be a python script, with a main() function. You should also randomly sample only 4000 rows to include in the csv.
    
\end{lstlisting}

\end{tcolorbox}

\section{Ablation Studies}
\label{sec:app-d}

We perform ablations on important design choices related to our prior. In addition to those reported in Section \ref{sec:results}, we experiment with a simplified prior design that uses a single MLP (\Cref{tab:ablation-different-prior}), but find that the 3-MLP design acheives greater performance, especially on the DPE metric.

Finally, instead of the histogram variant of the causal data-prior loss, we test a loss based on the continuous ranked probability score. In \Cref{tab:ablation-loss} we find that the CRPS loss gives slight improvement on TCGA, but otherwise is not as effective as the causal data-prior loss.

\begin{table*}[ht]
\centering
\caption{Ablation on the prior design. Ours is the 3-MLP prior discussed in Section \ref{sec:method}; 1-MLP is a single MLP prior. Reported metrics are mean MISE and mean DPE (lower is better).}
\label{tab:ablation-different-prior}
\small

\setlength{\tabcolsep}{2.5pt}
\setlength{\extrarowheight}{-1pt}
\setlength{\aboverulesep}{0.5ex}
\setlength{\belowrulesep}{0.5ex}
\setlength{\cmidrulesep}{0.3ex}

\begin{tabular}{lcccccccccccc}
\toprule

& \multicolumn{2}{c}{\makecell{Debt\\ $(\times 10^{-2})$}}
& \multicolumn{2}{c}{\makecell{MVICU \\ $(\times 10^{-3})$}}
& \multicolumn{2}{c}{{Warfarin}}
& \multicolumn{2}{c}{{TCGA}}
& \multicolumn{2}{c}{{News}}
& \multicolumn{2}{c}{\makecell{NewsHet\\ $(\times 10^{-2})$}} \\

\cmidrule(lr){2-3}
\cmidrule(lr){4-5}
\cmidrule(lr){6-7}
\cmidrule(lr){8-9}
\cmidrule(lr){10-11}
\cmidrule(lr){12-13}

\textbf{Method}
& \textbf{MISE}
& \textbf{DPE}
& \textbf{MISE}
& \textbf{DPE}
& \textbf{MISE}
& \textbf{DPE}
& \textbf{MISE}
& \textbf{DPE}
& \textbf{MISE}
& \textbf{DPE}
& \textbf{MISE}
& \textbf{DPE}
\\

\midrule

\textbf{Ours (3-MLP)}
& \textbf{2.22}
& \textbf{0.011}
& \textbf{1.45}
& --
& 36.6
& \textbf{2.62}
& 8.63
& 38.9
& 1.57
& \textbf{3.80}
& \textbf{5.58}
& \textbf{0.152}
\\

1-MLP
& 5.32
& 0.122
& 1.67
& --
& \textbf{27.2}
& 3.08
& \textbf{4.23}
& \textbf{37.8}
& \textbf{1.56}
& 4.37
& 5.75
& 0.250
\\

\bottomrule
\end{tabular}
\end{table*}

\begin{table*}[ht]
\centering
\caption{Ablation on choice of loss function (CRPS v.s. CE). Reported metrics are mean MISE and mean DPE (lower is better).}
\label{tab:ablation-loss}
\small

\setlength{\tabcolsep}{2.5pt}
\setlength{\extrarowheight}{-1pt}
\setlength{\aboverulesep}{0.5ex}
\setlength{\belowrulesep}{0.5ex}
\setlength{\cmidrulesep}{0.3ex}

\begin{tabular}{lcccccccccccc}
\toprule

& \multicolumn{2}{c}{\makecell{Debt\\ $(\times 10^{-2})$}}
& \multicolumn{2}{c}{\makecell{MVICU \\ $(\times 10^{-3})$}}
& \multicolumn{2}{c}{{Warfarin}}
& \multicolumn{2}{c}{{TCGA}}
& \multicolumn{2}{c}{{News}}
& \multicolumn{2}{c}{\makecell{NewsHet\\ $(\times 10^{-2})$}} \\

\cmidrule(lr){2-3}
\cmidrule(lr){4-5}
\cmidrule(lr){6-7}
\cmidrule(lr){8-9}
\cmidrule(lr){10-11}
\cmidrule(lr){12-13}

\textbf{Method}
& \textbf{MISE}
& \textbf{DPE}
& \textbf{MISE}
& \textbf{DPE}
& \textbf{MISE}
& \textbf{DPE}
& \textbf{MISE}
& \textbf{DPE}
& \textbf{MISE}
& \textbf{DPE}
& \textbf{MISE}
& \textbf{DPE}
\\

\midrule

\textbf{Ours (CE)}
& \textbf{2.22}
& \textbf{0.011}
& \textbf{1.45}
& --
& \textbf{36.6}
& \textbf{2.62}
& 8.63
& 38.9
& \textbf{1.57}
& \textbf{3.80}
& \textbf{5.58}
& \textbf{0.152}
\\

CRPS
& 3.06
& 0.039
& 1.57
& --
& 38.9
& 2.73
& \textbf{8.03}
& \textbf{37.2}
& 1.58
& \textbf{3.80}
& 5.90
& 0.450
\\

\bottomrule
\end{tabular}
\end{table*}

\end{document}